\documentclass[runningheads]{llncs}

\usepackage{eccv}

\usepackage{eccvabbrv}

\usepackage{graphicx}
\usepackage{booktabs}
\usepackage{bbm}
\usepackage{wrapfig}
\usepackage{multirow}
\usepackage{pdfpages}

\newcommand{\drop}[1]{\,\scriptsize{\color{gray}(#1)}}

\usepackage[accsupp]{axessibility}  
\usepackage{enumitem}
\setlist{topsep=2pt, itemsep=2pt, parsep=0pt, partopsep=0pt}

\usepackage{hyperref}

\usepackage{orcidlink}

\begin{document}

\title{From Hallucination to Grounding: Diagnosing Visual Spatial Intelligence via CRISP} 

\titlerunning{From Hallucination to Grounding}

\author{Zhixing Li\orcidlink{0009-0003-8014-9779} \and
Yinan Yu\orcidlink{0000-0002-3221-7517}}

\authorrunning{Z.~Li and Y.~Yu}

\institute{Chalmers University of Technology, Gothenburg, Sweden \\
\email{\{zhixingl, yinan\}@chalmers.se}}

\maketitle

\begin{abstract}
Current VLM evaluations often conflate language priors with genuine spatial reasoning. To address this, we introduce CRISP, a novel structural-diagnostic evaluation paradigm that assesses visual spatial intelligence through consistency, the alignment between implicit perception and explicit reasoning. Unlike traditional black-box QA, CRISP utilizes metric 3D Scene Graphs and an oracle intervention protocol to decouple latent reasoning capabilities from perceptual bottlenecks. This granular diagnosis uncovers a systematic perception-reasoning disconnect. Crucially, we reveal that while proprietary models possess robust latent reasoning engines, they suffer from inaccurate metric estimation and a critical failure to leverage their implicit structural representations. Conversely, open-source models remain fundamentally bottlenecked by their lack of multi-hop compositional reasoning. By shifting the focus from merely ``guessing correctly'' via language priors to genuinely ``perceiving, verifying, and reasoning,'' CRISP offers a rigorous roadmap for multimodal alignment beyond end-to-end post-training. The code and dataset are available at \url{https://github.com/iiyamayuki/CRISP-Bench}.

\end{abstract}

\section{Introduction}
\label{sec:intro}
Visual-spatial intelligence requires Vision-Language Models (VLMs) to go beyond fundamental perceptual capabilities regarding location and attributes, and instead comprehensively understand the relative spatial relationships among distinct objects \cite{yu2025far}. Building upon this understanding, VLMs are expected to accomplish complex reasoning tasks, such as spatial imagination and planning. This capability is a fundamental prerequisite for embodied AI and autonomous systems, where interacting with the physical world requires a leap from \textbf{static semantics} to \textbf{actionable spatial reasoning.} For instance, a robot placing a glass on a table must infer depth, estimate clearance, and reason about stable surfaces, which demands \textbf{implicit 3D spatial cognition} \cite{yang2025cambrian}.

Despite recent advancements that have equipped state-of-the-art VLMs with remarkable semantic recognition capabilities~\cite{achiam2023gpt, bai2023qwen, li2023blip, li2025latent, liu2023visual, radford2021learning, zhou2022learning}, a critical gap persists, rooted in the distinction between \textbf{atomic semantic perception} and \textbf{structural spatial perception}. While current models excel at the former: identifying ``what is where'' via semantic labels and 2D bounding boxes; they exhibit a fundamental deficit in the latter: the ability to mentally construct the ``intrinsic geometric structure'' of a scene, such as viewpoint transformations, relative distances, and spatial relationships~\cite{yang2025thinking}. Empirical studies suggest that even advanced VLMs often hallucinate spatial relationships or fail to maintain logical consistency when viewed from an embodied perspective \cite{ramakrishnan2024does, yu2025far, kamath-etal-2023-whats}. This disconnect indicates that while VLMs have mastered 2D semantic recognition, they have yet to grasp the implicit 3D spatial structure required in the physical world.

\begin{figure}[tb]
    \centering
    \includegraphics[width=0.75\linewidth]{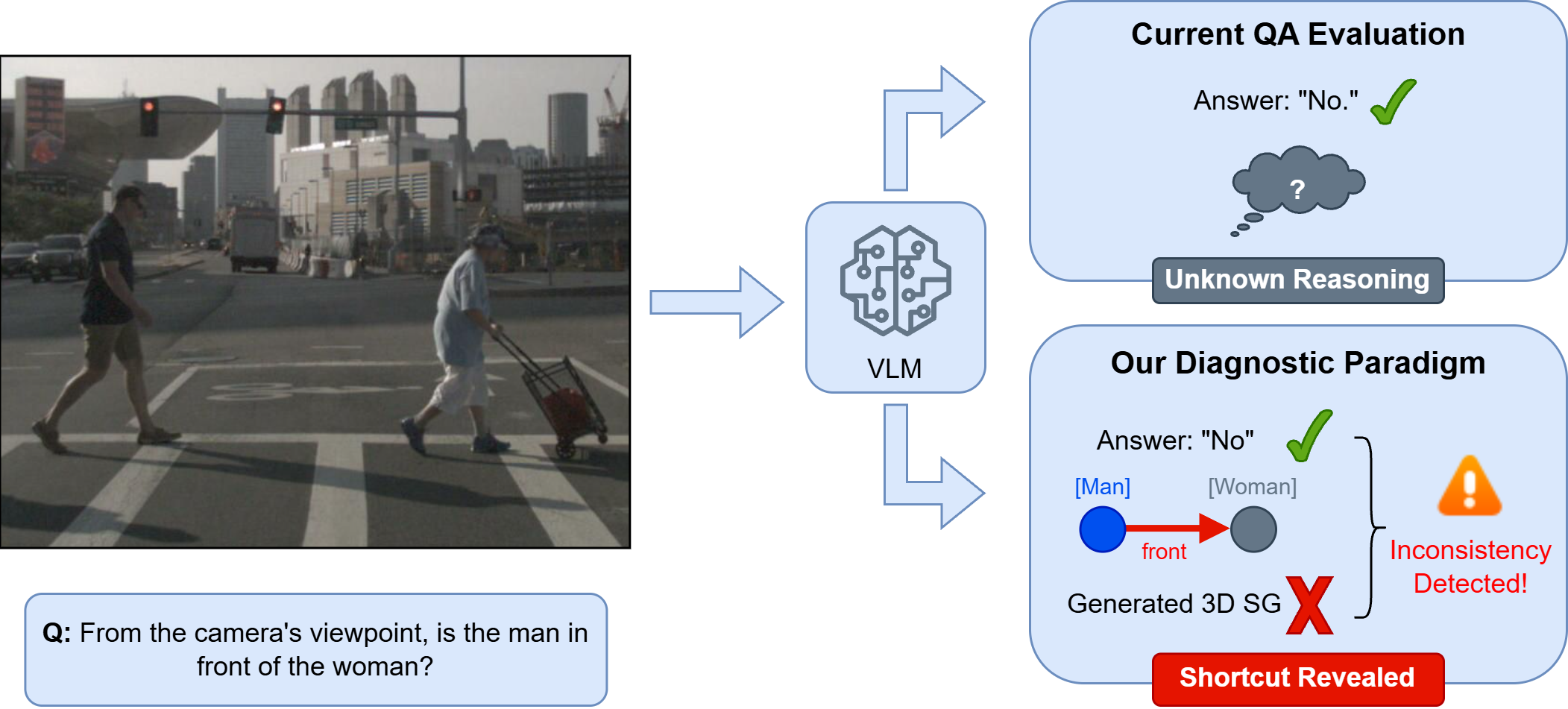}
    \caption{\textbf{The illusion of spatial intelligence.} While VLMs correctly answer queries, our structural-diagnostic paradigm reveals internal hallucinations. This exposes a critical inconsistency between correct linguistic outputs and flawed implicit 3D modeling.}
    \label{fig:intro}
\end{figure}

Precise evaluation is hampered by two limitations: traditional Question Answering (QA) formats allow models to exploit language priors for ungrounded correct answers \cite{10.5555/3737916.3738458,11148062}, while current spatial probes like textual rationales \cite{yang2025thinking} or 2D cognitive maps \cite{zhang2026theory} suffer from hallucinations or lack 3D metric precision. Moving beyond conventional VQA benchmarks, an effective probe must force models to explicitly reveal how they perceive the scene. We introduce a novel paradigm coupling QA with 3D Scene Graph (3D SG) \cite{armeni20193d} construction. Serving as a strict informational bottleneck, the 3D SG demands precise metric geometry and topology, definitively verifying whether reasoning is genuinely grounded or merely exploiting semantic shortcuts.

Our main contributions are summarized as follows:
\begin{enumerate}
    
\item \textbf{A Dual-Task Paradigm for Structural Spatial Diagnosis.} We introduce \textbf{CRISP} (\textbf{C}onsistency of \textbf{R}easoning \textbf{I}n \textbf{S}patial \textbf{P}erception), a benchmark of 1{,}162 static, single-view indoor/outdoor scenes with 9{,}839 questions. Focusing on static geometric grounding as the fundamental prerequisite for spatial intelligence, CRISP moves beyond traditional black-box QA evaluation. It couples discriminative QA with a generative 3D Scene Graph Construction (SGC) task, which compels VLMs to explicitly externalize their implicit spatial modeling. By concretizing abstract spatial intuition into actionable formats, CRISP shifts evaluation from \textit{result-oriented} correctness to \textit{process-oriented} grounding, offering a granular diagnosis of spatial intelligence that neither task affords in isolation.

\item \textbf{A Cross-Task Consistency Protocol for Grounding Verification.} We introduce a novel metric to evaluate the alignment between two fundamentally distinct output representations: the linguistic reasoning in QA and the structural topology in SGC. Unlike self-consistency methods~\cite{wang2022self} that merely check agreement within the same task, our protocol scrutinizes whether the model's textual answers are physically anchored in its generated geometry. This effectively distinguishes \textit{true visual reasoning} from \textit{semantic shortcuts}, validating that the answer is supported by the model's own perception.

   
\item \textbf{An Intervention-Based Diagnostic Protocol.} To explicitly disentangle perceptual bottlenecks from reasoning deficits, we design a dual-intervention experiment that contrasts model performance under ground-truth structural guidance versus self-predicted graphs. Crucially, this protocol uncovers a systematic \textit{Perception-Reasoning Disconnect:} state-of-the-art VLMs possess robust latent reasoning capabilities but fail to anchor them in their own visual perception. This finding redefines the development roadmap, identifying structural alignment rather than intrinsic logical deficiency as the primary bottleneck for future embodied models.

\end{enumerate}

\section{Related Works}
\label{sec:related_works}

\textbf{Visual Spatial Intelligence.} 
Visual spatial intelligence encompasses perception, reasoning, and linguistic interaction~\cite{yang2025thinking}. A critical component is implicit 3D spatial cognition, which interprets 2D images as projections of a 3D world and is identified as essential for predictive world modeling~\cite{yang2025cambrian}.
Current advancements in this domain can be categorized into three streams: \textbf{(1) auxiliary geometric priors:} To bridge the 2D-3D gap, SpatialRGPT \cite{cheng2024spatialrgpt} and SpatialCLIP \cite{wang2025spatialclip} explicitly inject depth maps, while VG LLM \cite{zheng2025learning} and SpatialMLLM \cite{wu2025spatial} leverage external encoders like VGGT \cite{wang2025vggt} to extract spatial features. \textbf{(2) Optimization strategies:} Beyond architecture, 3D-VisTA \cite{zhu20233d} and TIPS \cite{kokitsi2025tips} enhance spatial understanding through refined pre-training objectives, while reinforcement learning is employed by SpaceR \cite{ouyang2025spacer} and SpatialReasoner \cite{ma2025spatialreasoner} to bolster logical reasoning. \textbf{(3) Inference prompting:} Focusing on the inference stage, VoT \cite{wu2024mind} and MVoT \cite{li2025imagine} adapt Chain-of-Thought (CoT) \cite{wei2022chain} to multimodal contexts, guiding models to decompose complex spatial queries.

\noindent \textbf{Spatial Reasoning Benchmarks.} Existing evaluations range from abstract cognitive probes on synthetic images (e.g., SPACE \cite{ramakrishnan2024does}, BSA \cite{xu-etal-2025-defining}, SpatialEval \cite{wang2024picture}) to realistic embodied benchmarks in navigable 3D spaces (e.g., 3DSR Bench \cite{ma20253dsrbench}, Open3D-VQA \cite{zhang2025open3dvqa}, SQA3D \cite{ma2022sqa3d}, Whatsup \cite{kamath-etal-2023-whats}, OmniSpatial \cite{jia2025omnispatial}, MMSI-bench \cite{yang2025mmsi}). Despite this domain divergence, these benchmarks predominantly rely on discriminative QA. Recent diagnostic works \cite{tong2024eyes, brown2025benchmark} reveal that such QA formats frequently allow VLMs to exploit ``semantic shortcuts'' rather than exhibiting genuine visual grounding across perceptual and spatial tasks. CRISP explicitly advances this diagnostic lineage by proposing a generative metric bottleneck to rigorously verify implicit 3D cognition. Furthermore, while efforts like SIGBench \cite{wu2025towards} and THEORY OF SPACE \cite{zhang2026theory} attempt to probe cognitive processes via macroscopic cognitive maps \cite{tolman1948cognitive}, transitioning from abstract cognitive probing to actionable embodied control requires a shift towards the precise, metric-aware representations introduced in CRISP.

\noindent \textbf{3D Scene Graphs.} Unlike traditional 2D scene graphs \cite{Johnson_2015_CVPR}, which primarily capture semantic adjacencies in the image plane, 3D SGs explicitly encode metric geometry, object pose, and spatial topology. This representation serves as a structured abstraction of the physical world, grounding semantic concepts into actionable geometric coordinates. Due to this nature, 3D SGs have emerged as a cornerstone in embodied AI and robotics. For instance, CURB-SG \cite{10610112} showcases the construction of dynamic 3D SGs within large-scale urban environments. ConceptGraphs \cite{gu2024conceptgraphs} illustrates how open-vocabulary 3D SGs can effectively bolster robotic perception and planning capabilities. Furthermore, SayPlan \cite{rana2023sayplan} demonstrates that grounding Large Language Models (LLMs) in 3D SGs significantly enhances the execution of complex tasks. 

\section{The CRISP Benchmark}
\label{sec:crisp}
\begin{figure}[tb]
    \centering
    \includegraphics[width=0.75\linewidth]{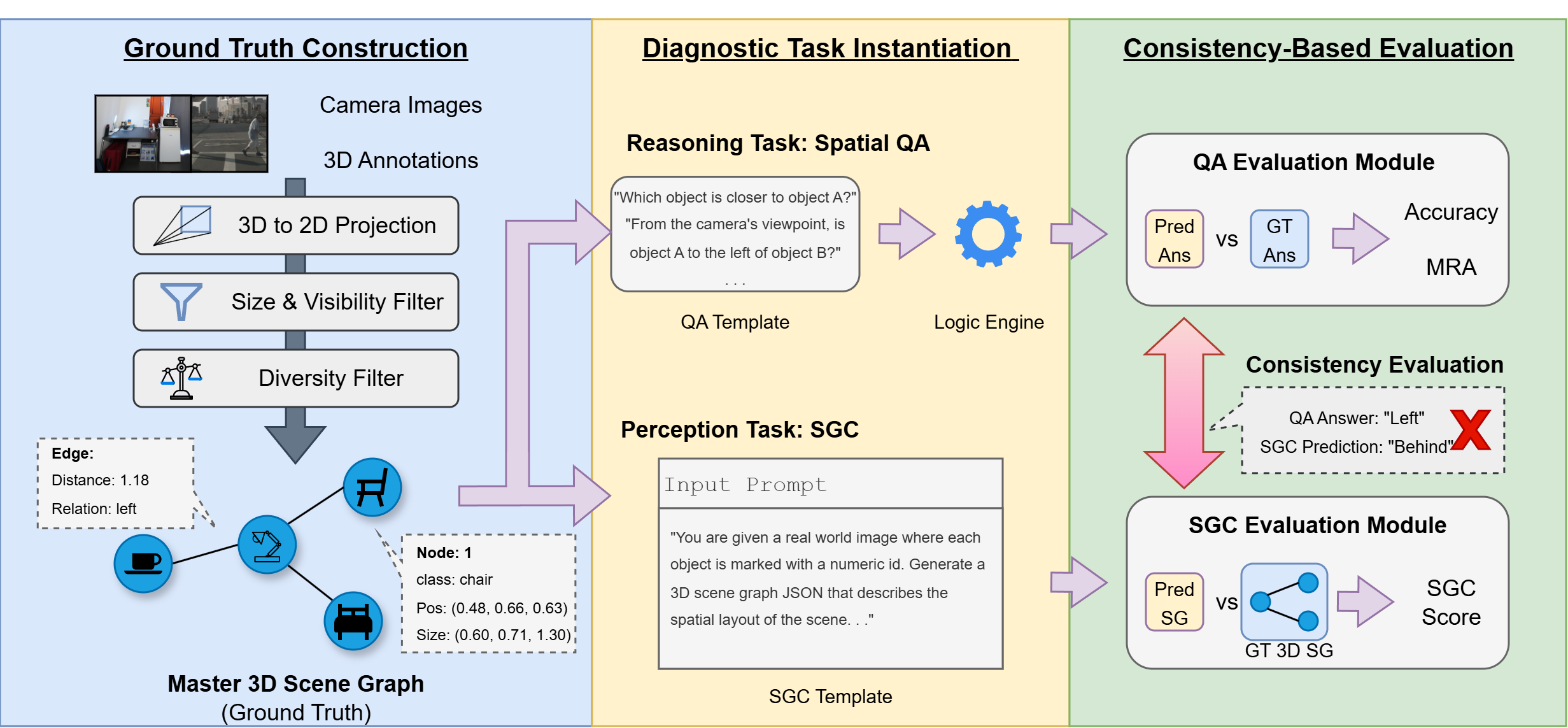}
    \caption{\textbf{Overview of the CRISP benchmark.} We construct a physically grounded 3D Scene Graph to instantiate paired diagnostic tasks: Spatial QA and SGC. A novel consistency protocol then evaluates the alignment between explicit linguistic reasoning and the externalization of implicit structural modeling.}
    \label{fig:framework}
\end{figure}
\subsection{Automated 3D Ground Truth Construction}
\label{sec:pipeline}
To construct a physically grounded benchmark, we source data from nuScenes \cite{caesar2020nuscenes} and ScanNet++ \cite{yeshwanth2023scannet++}. NuScenes provides macro-scale, dynamic outdoor driving environments, while ScanNet++ offers micro-scale, high-fidelity indoor scenes. In total, we curated a set of 1,162 high-quality master scenes, maintaining a strict 1:1 ratio between indoor and outdoor domains. Rather than pursuing massive but noisy data scaling, this rigorously balanced dataset provides sufficient statistical power for fine-grained diagnostic evaluation while strictly eliminating semantic redundancy. Regarding the temporal dimension, we prioritize static imagery over video sequences. This choice is grounded in the premise that static spatial perception serves as a prerequisite for spatiotemporal comprehension. By isolating spatial reasoning from temporal dynamics (e.g., motion blur), we ensure a focused evaluation of the VLMs' structural scene reasoning.

As illustrated in the Ground Truth Construction module of \cref{fig:framework}, we implement a rigorous automated preprocessing pipeline. Initially, we leverage sensor calibration matrices to project high-quality 3D annotations onto the 2D image plane, aligning 3D primitives with 2D observations. To ensure unambiguous visual grounding, we apply a strict visibility filter that prunes objects subject to heavy occlusion (e.g., visibility score < 0.8), truncation, or insufficient pixel resolution (e.g., 2D bounding box longest edge < 40 pixels). Furthermore, a diversity filter is employed to mitigate semantic redundancy by removing samples with highly similar 3D SGs, balancing the dataset distribution based on scene complexity (detailed in Appendix A). Finally, for each curated sample, we construct a Master 3D Scene Graph utilizing unique numerical IDs \cite{yang2023set}. This design circumvents the semantic leakage of textual descriptions and the geometric priors of 2D bounding boxes, providing an unbiased reference for grounding.

\subsection{Diagnostic Task Design}
\begin{wrapfigure}{r}{0.5\textwidth}
    \centering
    \includegraphics[width=\linewidth]{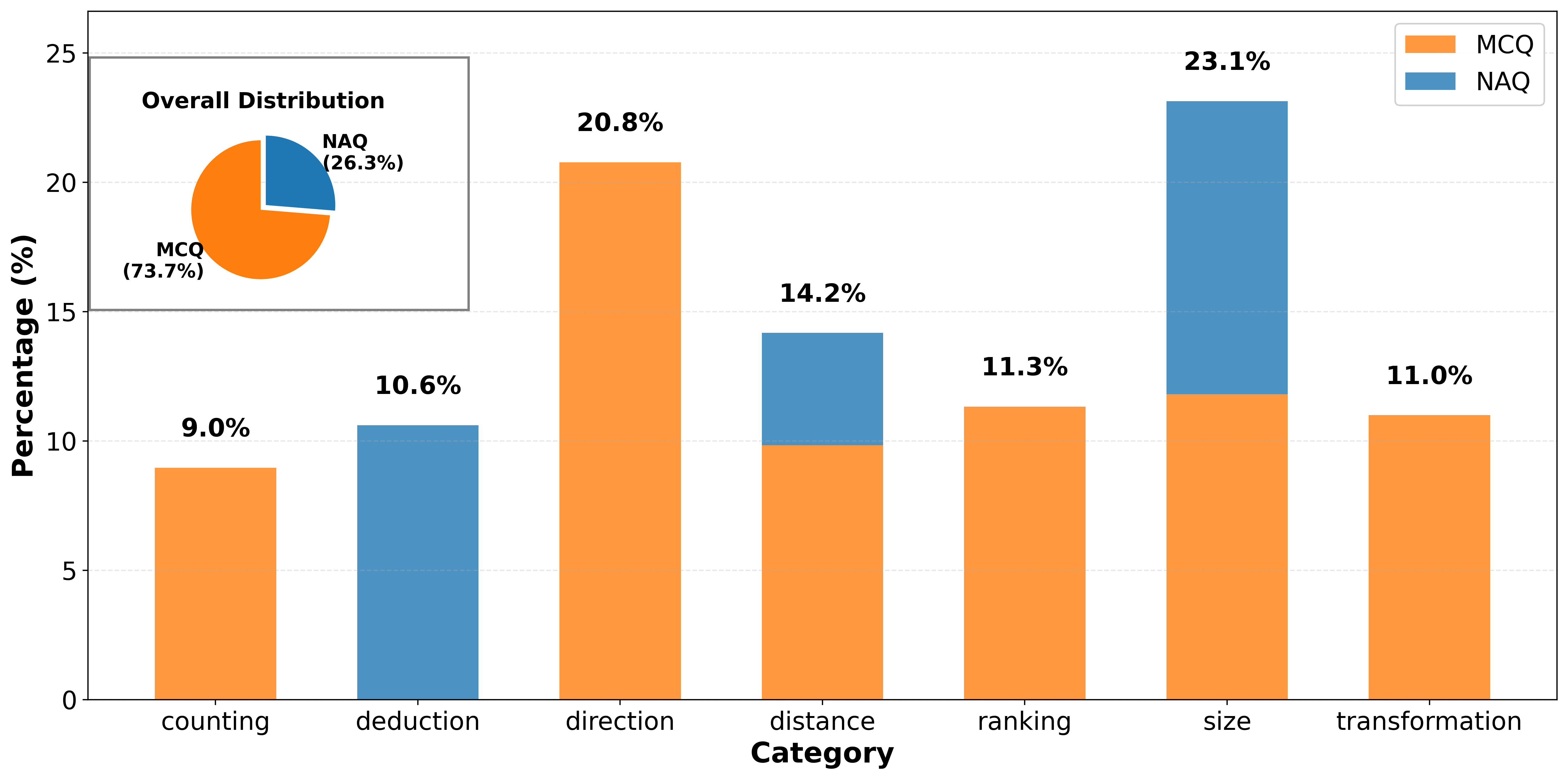}
    \caption{Question distribution of the Spatial QA task.}
    \label{fig:qa_stat}
\end{wrapfigure}

\textbf{Spatial QA Task.} To probe explicit reasoning, we instantiate a Spatial QA suite comprising seven competencies vital for embodied agents: \textit{distance estimation} (reaching), \textit{size estimation} (clearance), \textit{directional perception} (egocentric navigation), \textit{counting} (inventory), \textit{spatial ranking} (prioritization), \textit{view transformation} (pose prediction), and \textit{logical deduction} (multi-hop inference). We deliberately exclude queries like containment or occlusion, as they inherently entangle with object-specific semantic priors rather than the pure geometric structures our benchmark targets.

Beyond traditional Multiple-Choice Questions (MCQs), we incorporate Numerical Answer Questions (NAQs), challenging models to perform precise quantitative estimation rather than simple classification. To ensure zero ambiguity, we developed a deterministic logic engine that procedurally generates image-specific questions by querying the Master 3D Scene Graph via hand-crafted templates (see Appendix A.5). The final dataset consists of 9,839 QA pairs with a rigorously balanced answer distribution for MCQs, as detailed in \cref{fig:qa_stat}.

\noindent\textbf{3D Scene Graph Construction (SGC) Task.} 
To scrutinize the intermediate spatial representations often obscured in end-to-end QA, we introduce the SGC task, forcing VLMs to formulate their implicit 3D modeling into structured graphs, making it as a direct probe for structural perception. In this task, we provide the model with a list of objects of interest. This protocol deliberately disentangles spatial reasoning from object discovery, ensuring that the evaluation targets the model's ability to infer relationships among known entities rather than its detection recall.  Furthermore, considering the prohibitive complexity of fully connected graphs for VLMs, we adopt an \textit{Object-Centric} generation strategy. Given a center object, the model generates a star-topology subgraph. This approach serves as a minimal yet sufficient proxy for local topology while mitigating the generation drift common in long-context outputs. As validated in \cref{sec:ablation}, this format ensures near-perfect syntactic compliance, effectively isolating structural perception from formatting errors.

Unlike human qualitative intuition, embodied control requires rigorous metric precision. While monocular metric estimation without camera intrinsics is mathematically ill-posed, we intentionally withhold them to test if VLMs can leverage world knowledge priors (e.g., physical object scales) as visual anchors to recover metrics. We strictly distinguish this necessary physical grounding from penalized linguistic shortcuts (blind textual guessing). Hence, our SGC task demands a hybrid ego-centric output:
\begin{itemize}
    \item \textbf{Nodes (Intrinsics):} The model predicts the 3D dimensions and distance to camera for each object.
    \item \textbf{Edges (Relations):} The output include both semantic directional predicates (e.g., \textit{front, left}) and the metric Euclidean distance between objects.
\end{itemize}

Consequently, SGC transcends standard visual grounding by evaluating the model's capability to articulate complex spatial relationships. While models may possess implicit spatial priors, this task assesses whether such knowledge can be reliably \textit{externalized} into the structured, metric-aware formats required for downstream planning and manipulation.

\subsection{Evaluation Metrics}
\label{sec:metrics}
\textbf{QA Score.} For the spatial QA tasks, we adopt evaluation metrics identical to those employed in VSI-bench \cite{yang2025thinking}. Specifically, for MCQs, we utilize \textit{Accuracy} based on exact matching; whereas for NAQs, we employ the \textit{Mean Relative Accuracy (MRA)}, defined as: $MRA=\frac{1}{10}\sum_{\theta\in \mathcal{C}}\mathbbm{1}\left(|\hat{y}-y|/y<1-\theta \right),$ where $\hat{y}$ is the model's prediction, $y$ is the ground truth, $\mathcal{C}=\{0.5,0.55,\cdots,0.95\}$ is the confidence thresholds. The final QA score is calculated as the macro-average of the accuracy or MRA across all question categories.

\noindent\textbf{SGC Score.} To quantify the fidelity of the externalized spatial representation, we design a composite metric that harmonizes \textit{geometric precision} with \textit{semantic correctness}. The SGC Score comprises two pillars:

\noindent\textit{(1) Metric Estimation ($S_{est}$).} Prioritizing physical plausibility, this component aggregates three geometric sub-scores.
\textit{Object Size Score ($S_{size}$)} evaluates dimensional accuracy using a ratio-based IoU formulation:
\begin{equation}
    S_{size} = \frac{1}{N}\sum_{i=1}^{N}\frac{1}{3} \sum_{k \in \{w, l, h\}} \frac{\min(k_p, k_g)}{\max(k_p, k_g)},
\end{equation}
where subscripts $p, g$ indicate prediction and ground truth. Crucially, this ratio-based design ensures scale invariance, treating errors on small and large objects equally.
For \textit{Distance Scores} (ego-centric $S_{dist\_to\_cam}$ and object-centric $S_{dist}$), we adopt a relative accuracy protocol to account for depth uncertainty:
\begin{equation}
    S_{dist\_to\_cam}, S_{dist} = \frac{1}{N}\sum_{i=1}^{N}\max \left( 0, 1 - \frac{|d_p - d_g|}{d_g + \epsilon} \right),
\end{equation}
where $\epsilon$ is a small number ($10^{-6}$). This penalizes absolute errors strictly in the near-field while allowing proportional margins in the far-field. The aggregate metric score is the mean of these three: $S_{est}=(S_{size}+S_{dist\_to\_cam}+S_{dist})/3$.

\noindent\textit{(2) Relation Score ($S_{rel}$).} Complementing metric precision, this component evaluates the semantic consistency of edge predicates. To rigorously penalize contradictory predictions (e.g., simultaneously predicting ``left'' and ``right''), we evaluate relations as mutually exclusive pairs (e.g., \{left, right\}, \{front, behind\}). This score measures the accuracy of these pair-wise predictions:
\begin{equation}
    S_{rel}=\frac{1}{N_{edge}}\sum_{i=1}^{N_{edge}}\frac{N_\text{Correct Pairs}}{N_\text{Total Pairs}}.
\end{equation}
Finally, $\textit{SGC Score}=(S_{est} + S_{rel}) / 2$, ensuring that high performance requires both rigorous metric grounding and logical semantic reasoning.

\noindent\textbf{Self-Consistency Score.} We posit that true visual spatial intelligence necessitates \textbf{internal coherence}. A model possessing a robust implicit 3D spatial cognition must yield non-contradictory outputs across different task modalities targeting the \textit{same} scene. Discrepancies between explicit reasoning (QA) and the externalization of implicit structural modeling (SGC) strongly suggest that the model is relying on linguistic shortcuts or superficial pattern matching rather than grounded understanding. Thus, we introduce the Self-Consistency Score to quantify this cross-task alignment. Unlike standard accuracy metrics, this score evaluates the model's self-consistency via a two-step protocol:
\begin{enumerate}
    \item \textbf{Symbolic Derivation:} We first parse the model's generated SGC output into a structured graph and employ a deterministic, rule-based solver to infer the corresponding QA answer. If the graph lacks necessary nodes or edges to support the inference, a \texttt{[FAILED]} token is assigned. At this stage, we can evaluate the derived QA answers ($A_{derived}$) using the ground truth, termed the \textit{Derived QA Score.}
    \item \textbf{Consistency Check:} We treat the model's direct QA response ($A_{QA}$) as the reference anchor. The consistency score is then computed by evaluating $A_{derived}$ against $A_{QA}$ using standard QA Score.
\end{enumerate}

This metric measures whether the model's QA response and its generated scene graph agree with \emph{each other}, not whether either is correct, since a model can be consistently wrong (termed ``consistent hallucination''). Genuine spatial intelligence requires joint verification across all three metrics.
\section{Experiments}
\label{sec:experiments}
\textbf{Evaluation Setup.} We benchmark 13 state-of-the-art VLMs, comprising proprietary giants like Gemini 2.5 Flash/Pro \cite{comanici2025gemini}, Gemini 3 Flash \cite{google_gemini3_2025}, GPT-5-Mini \cite{singh2025openai} and GPT-5.2 \cite{singh2025openai}; and leading open-source models like Qwen2.5/3-VL \cite{bai2025qwen2, bai2025qwen3vltechnicalreport}, InternVL-3.5 \cite{wang2025internvl3}, LLaVA-OneVision-1.5 \cite{an2025llava}. We also include VG LLM \cite{zheng2025learning} and Cambrian-S \cite{yang2025cambrian} as specialized spatial baselines. All evaluations utilize the lmms-eval \cite{zhang2025lmms} framework in a zero-shot setting. To ensure fair comparison, we disable the ``thinking'' mode for most proprietary models. For models where this feature is mandatory, we enforce a low reasoning budget ($\sim$1,024 tokens). This protocol mitigates the confounding effects of test-time compute, isolating the models' intrinsic visual spatial perception capabilities from text-driven self-correction. Default system prompts and greedy decoding are employed for reproducibility.

\subsection{Main Results: The State of Visual Spatial Intelligence}
\label{sec:main_results}
\begin{figure*}[t]
    \centering
    \begin{minipage}[c]{0.58\linewidth} 
        \centering
        \captionof{table}{Main evaluation results. We report the aggregate QA, SGC (includes $S_{est}$ and $S_{rel}$) and Consistency (Con.) scores. \textbf{Bold} denotes the best performance within each category.}
        \label{tab:main_results}
        
        \resizebox{\linewidth}{!}{
            \setlength{\tabcolsep}{3.5pt} 
            \begin{tabular}{l c c c c c}
                \toprule
                \textbf{Model} & \textbf{QA} & $\mathbf{S_{est}}$ & $\mathbf{S_{rel}}$ & \textbf{SGC} & $\mathbf{Con.}$ \\
                \midrule
                \multicolumn{5}{l}{\textit{\textbf{Proprietary Models}}} \\ 
                Gemini-2.5-Flash & 47.68 & 61.24 & 68.61 & 64.92 & 49.98 \\
                Gemini-2.5-Pro$^\dagger$ & \textbf{58.93} & 57.79 & 71.36 & 64.58 & \textbf{57.28} \\
                Gemini-3-Flash & 53.41 & 65.25 & \textbf{72.33} & \textbf{68.79} & 55.20 \\
                GPT-5-Mini$^\dagger$ & 54.15 & 61.39 & 72.01 & 66.70 & 56.75 \\
                GPT-5.2 & 48.75 & \textbf{66.67} & 64.61 & 65.64 & 48.87 \\
                \midrule
                \multicolumn{5}{l}{\textit{\textbf{Open-source Models}}} \\
                LLaVA-OV-1.5-8B & 47.85 & 44.53 & 56.12 & 50.33 & 38.98 \\
                InternVL3.5-8B & 48.37 & 51.32 & 61.10 & 56.21 & 43.59 \\
                InternVL3.5-38B & 53.98 & 53.12 & 64.68 & 58.90 & 48.70 \\
                Qwen2.5-VL-7B & 44.78 & 40.62 & 44.79 & 42.70 & 35.62 \\
                Qwen3-VL-8B & \textbf{55.34} & 57.09 & \textbf{67.62} & 62.35 & \textbf{52.16} \\
                Qwen3-VL-32B & 53.98 & \textbf{60.21} & 64.69 & \textbf{62.45} & 51.56 \\
                \midrule
                \multicolumn{5}{l}{\textit{\textbf{Specialized Baselines}}} \\
                Cambrian-S & \textbf{45.95} & \textbf{46.76} & 47.50 & 47.13 & 34.22 \\
                VG LLM & 38.34 & 40.61 & \textbf{53.85} & \textbf{47.23} & \textbf{35.29} \\
                \bottomrule
            \end{tabular}
        }
        \par
        \scriptsize{\raggedright $^\dagger$: Thinking mode restricted.\par}
    \end{minipage}%
    \hfill
    \begin{minipage}[c]{0.40\linewidth}
        \centering
        \includegraphics[width=0.9\linewidth]{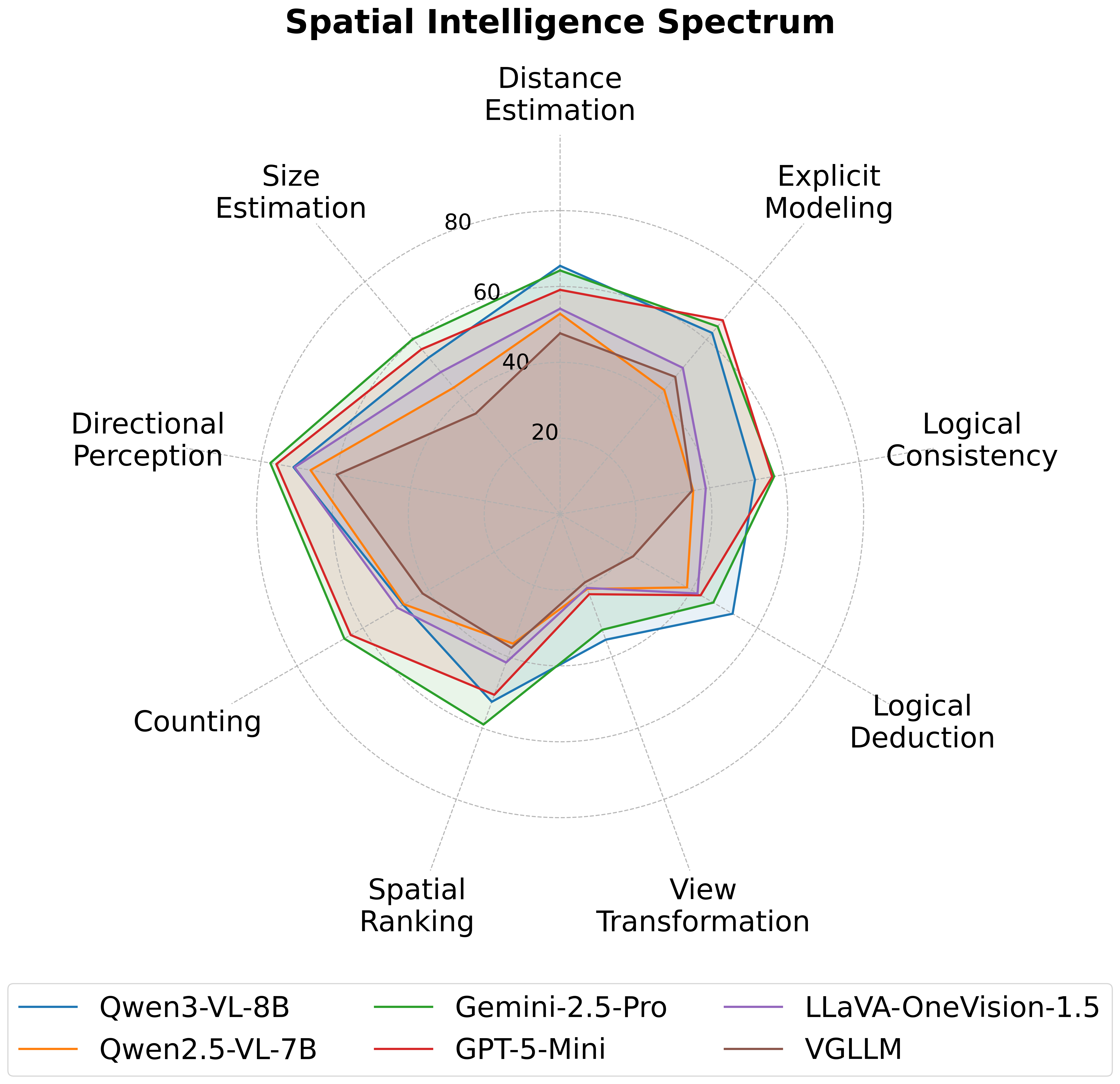}
        \caption{Fine-grained spatial capabilities breakdown across 9 dimensions.}
        \label{fig:radar_plot}
    \end{minipage}
\end{figure*}

\begin{figure*}[t]
    \centering
        \begin{subfigure}[t]{0.48\linewidth}
        \centering
        \includegraphics[width=\linewidth]{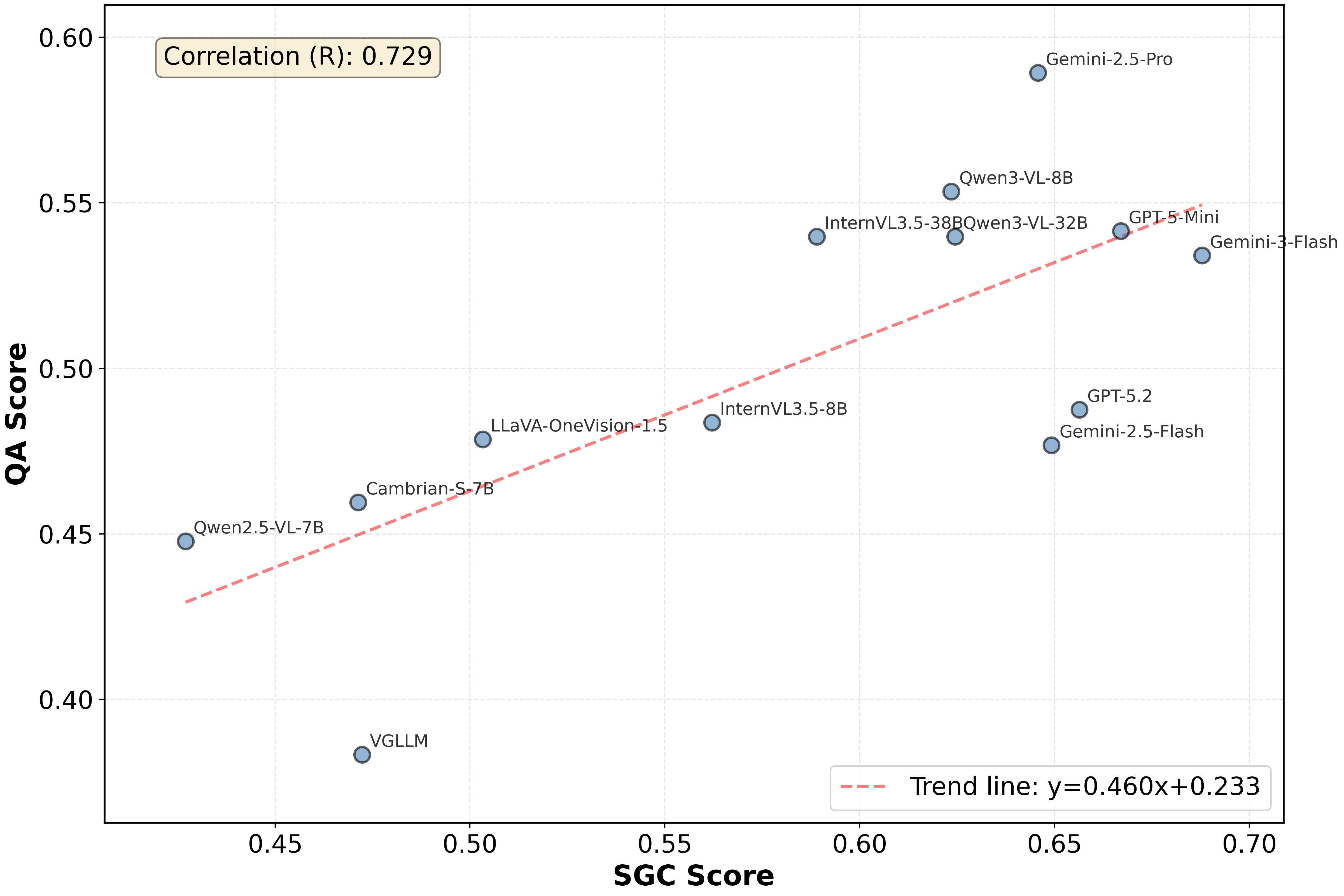}
        \caption{Perception-Reasoning Alignment.}
        \label{fig:qa_sgc}
    \end{subfigure}
    \hfill
    \begin{subfigure}[t]{0.48\linewidth}
        \centering
        \includegraphics[width=\linewidth]{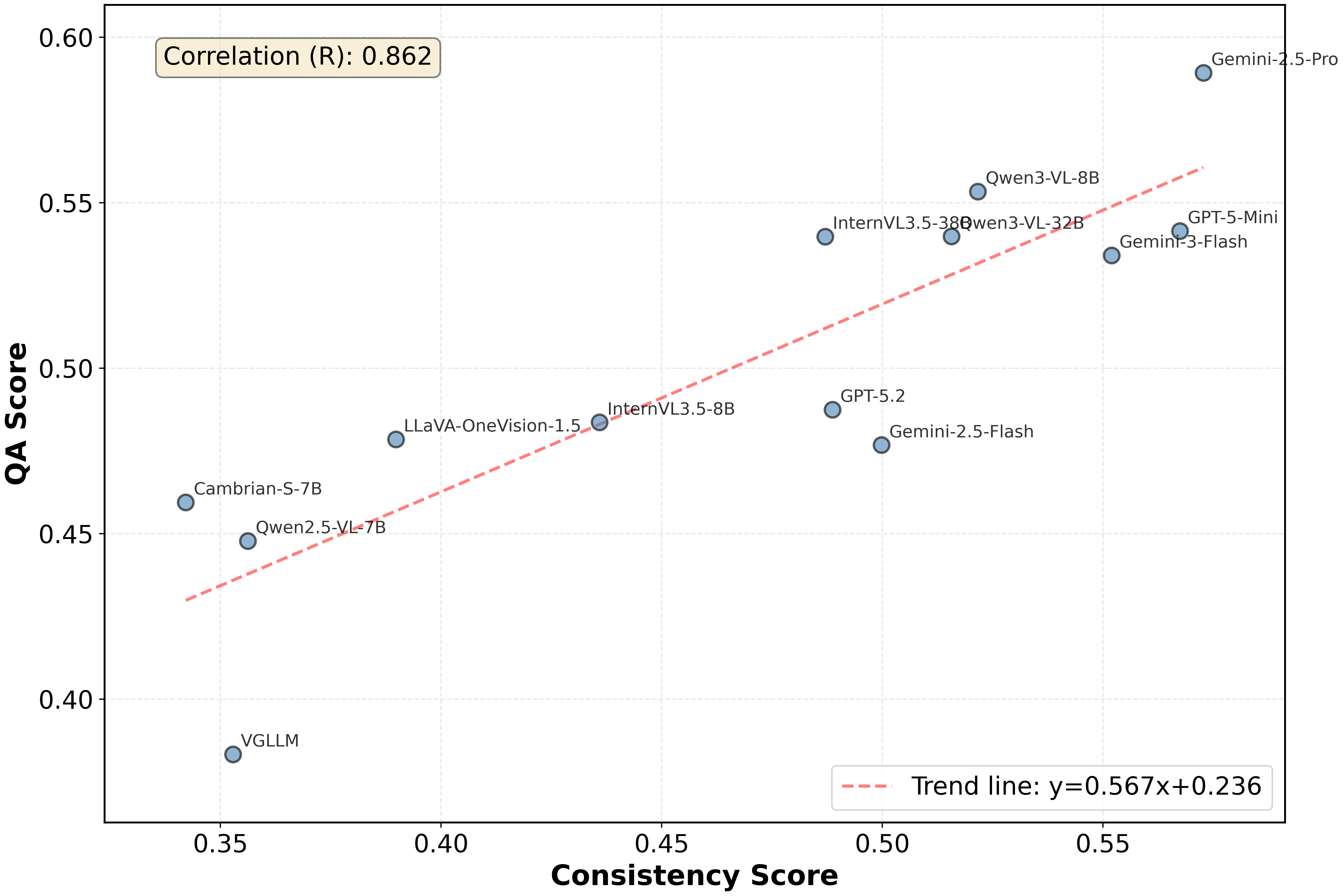}
        \caption{Reliability Verification via Consistency.}
        \label{fig:qa_consistency}
    \end{subfigure}
    
    \caption{\textbf{Visualizing the dependency of reasoning on grounding.} Explicit reasoning (QA) is fundamentally bottlenecked by the fidelity of implicit structural perception (SGC), while consistency serves as the critical indicator of genuine understanding.}
    \label{fig:scatter_analysis}
\end{figure*}

\textbf{Spatial QA.} Table \cref{tab:main_results} highlights a shrinking gap between proprietary and open-source models, led by Gemini 2.5 Pro and Qwen3-VL-8B respectively. While models excel at fundamental tasks (e.g., Counting, Direction) as \cref{fig:radar_plot}, they struggle with mental simulation (View Transformation). Notably, QA performance does not strictly scale with parameter size, suggesting that data quality and architectural priors outweigh raw scale for spatial intelligence. Finally, specialized spatial models surprisingly underperform generalist baselines, a counter-intuitive finding we will discuss later.

\noindent\textbf{SGC.} As illustrated in \cref{tab:main_results}, decomposing SGC reveals that models perform better on topological relations ($S_{rel}$) than metric estimation ($S_{est}$). Proprietary models exhibit moderate ability to utilize physical priors to recover scene scale, whereas open-source models score lower, indicating a lack of metric grounding. Interestingly, specialized models like VG LLM show marginal SGC gains compare to Qwen-2.5-VL yet fail at QA, hinting at fragmented capabilities.

Comparing SGC and QA \cref{fig:qa_sgc} reveals a clear utilization failure. While Gemini 2.5 Flash matches the Pro version in grounding ($\Delta$SGC=+0.34), it trails significantly in QA task ($\Delta$QA=-11.25), indicating it builds reasonable structural models but fails to leverage them during reasoning. Furthermore, a high QA score combined with a moderate SGC score creates an interpretability challenge: it could imply efficient perception-reasoning conversion, but equally suggests a reliance on language priors. Since aggregate metrics cannot disentangle these two possibilities, we cannot rely on the score disparity alone. Instead, we must examine the instance-level alignment to verify if the reasoning is genuinely grounded. This leads us to the core findings in our \textit{Consistency} Analysis.

\noindent\textbf{Consistency.} As detailed in Appendix B.3, proprietary models often achieve Derived QA $>$ Base QA. This indicates the \textit{perception-reasoning disconnect:} their structural perception is adequate but their reasoning engines fail to leverage it during end-to-end generation. Open-source models show the opposite: correct QA answers without corresponding structural grounding, quantifying hallucination and shortcut reliance.

The Consistency Score serves as a prerequisite for validating structural alignment. For instance, despite outperforming Gemini 2.5 Flash in QA, LLaVA-OneVision-1.5's low consistency exposes its reliance on semantic shortcuts. Conversely, the Qwen3-VL series maintains high coherence, validating reliable spatial reasoning. Empirically, the ``consistent hallucination'' phenomenon introduced in \cref{sec:metrics} occurs in approximately 10-12\% of evaluated instances (detailed via the confusion matrix in Appendix B.4). This bounded subset indicates systematic perceptual failures or the strong influence of intrinsic language priors, where models reason coherently but over fundamentally flawed geometric grounding. Notably, the low coherent errors in Gemini models (e.g., ~3.8\%) serves as a signature of their perception-reasoning disconnect. Furthermore, the off-diagonal elements of the matrix visually corroborate our core findings: the prominent volume of shortcut-driven successes and disconnect-driven failures.

Finally, specialized models highlight potential limitations in spatial post-training. While format mismatch might be suspected, Cambrian-S and VG~LLM actually outperform standard baselines (e.g., Qwen2.5-VL) on $S_{est}$ and $S_{rel}$ yet exhibit lower consistency, suggesting a fragmented cross-task alignment. Since post-training typically elicits pre-existing capabilities rather than injecting new cognition~\cite{zhou2023lima, yue2025does}, spatial fine-tuning may activate local structural cues but fails to ensure the reasoning engine utilizes them. Consequently, ungrounded QA optimization may encourage semantic shortcuts over genuine spatial intelligence.

\subsection{Ablation Study}
\label{sec:ablation}
\begin{wraptable}{htbp}{0.48\textwidth}
    \centering
    \caption{\textbf{Blind Test Analysis.} We report the impact of removing visual input. Values in parentheses denote the performance drop. Specialized models are excluded due to lack of text-only support.}
    \label{tab:blind_test}
    
    \resizebox{\linewidth}{!}{
        \setlength{\tabcolsep}{3pt}
        \begin{tabular}{l c c}
            \toprule
            \textbf{Model} & \textbf{QA} & \textbf{SGC} \\
            \midrule
            \multicolumn{3}{l}{\textit{\textbf{Proprietary Models}$^*$}} \\ 
            Gemini-2.5-Flash     & 28.02\drop{-19.66} & 45.33\drop{-19.59} \\
            Gemini-2.5-Pro & 26.90\drop{-32.03} & 37.05\drop{-27.53} \\
            Gemini-3-Flash       & 7.94\drop{-45.47}  & 39.47\drop{-29.32} \\
            GPT-5-Mini     & 18.31\drop{-35.84} & 39.32\drop{-27.38} \\
            GPT-5.2              & 30.34\drop{-18.41} & 38.39\drop{-27.25} \\
            \midrule
            \multicolumn{3}{l}{\textit{\textbf{Open-source Models}}} \\
            LLaVA-OV-1.5-8B      & 22.39\drop{-25.46} & 43.20\drop{-7.13} \\
            InternVL3.5-8B       & 32.31\drop{-16.06} & 41.45\drop{-14.76} \\
            InternVL3.5-38B      & 33.78\drop{-20.20} & 41.66\drop{-17.24} \\
            Qwen2.5-VL-7B        & 28.47\drop{-16.31} & 39.46\drop{-3.24} \\
            Qwen3-VL-8B          & 31.29\drop{-24.05} & 34.75\drop{-27.60} \\
            Qwen3-VL-32B         & 32.78\drop{-21.20} & 35.56\drop{-26.89} \\
            \midrule
            Random baseline & 25.85 & 28.13 \\
            \bottomrule
        \end{tabular}
    }
    \par
    \scriptsize{\raggedright $^*$: Affected by text-only refusal behaviors. See \cref{sec:ablation} for details.\par}
\end{wraptable}
To isolate genuine structural perception from dataset biases, we conducted a blind test (text-only input). Crucially, since the prompts references objects solely by numerical IDs without semantic labels (as \cref{sec:pipeline}), the model is strictly prevented from exploiting semantic co-occurrence priors. As shown in \cref{tab:blind_test}, text-only models generally establish a statistical floor around $\sim40$ SGC Score. Notably, models like Gemini 3 Flash exhibit severe performance drops in QA. This is driven by conservative safety alignments triggering refusal behaviors when visual input is absent (e.g., explicitly declining to guess spatial relations). This performance reflects the models' ability to leverage distributional priors rather than semantic shortcuts: generating syntactically valid graphs and estimating metrics within statistically plausible ranges. This establishes a rigorous baseline where any performance gain must stem from actual visual signal processing.

The addition of visual input reveals the deceptive nature of visual gain in standard QA. While proprietary models achieve substantial visual gains on the SGC task, demonstrating a genuine ability to override distributional guesses with precise perception, open-source models like Qwen2.5-VL expose the illusion of semantic grounding. Specifically, while Qwen2.5-VL's QA score improves significantly with visual input ($\Delta \text{QA}=16.31$), its SGC gain is very limited ($\Delta \text{SGC}=3.24$). This dissociation implies a critical \textit{Semantic-Geometric Gap}: the vision encoder successfully identifies what the objects are, allowing the logical engine to answer QA queries based on restored semantic contexts, but fails to translate this recognition into where they are located. Consequently, the SGC output falls back to the text-only distributional prior. This strongly indicates that standard QA benchmarks often conflate semantic recognition with spatial understanding, whereas CRISP successfully diagnoses this nuance.

Finally, this ablation isolates the source of SGC failures. In the text-only setting, despite generating hallucinatory content, all models maintained a near-perfect 3D SG format compliance rate ($>99\%$). It shows that models are fully capable of generating complex, syntactically correct scene graphs. Consequently, the low SGC scores observed in \cref{tab:main_results} stem fundamentally from a lack of visual spatial perception, not a lack of instruction-following capability.

\subsection{Qualitative Analysis}
\label{sec:case_study}
\begin{figure}[tb]
    \centering
    \includegraphics[width=0.75\linewidth]{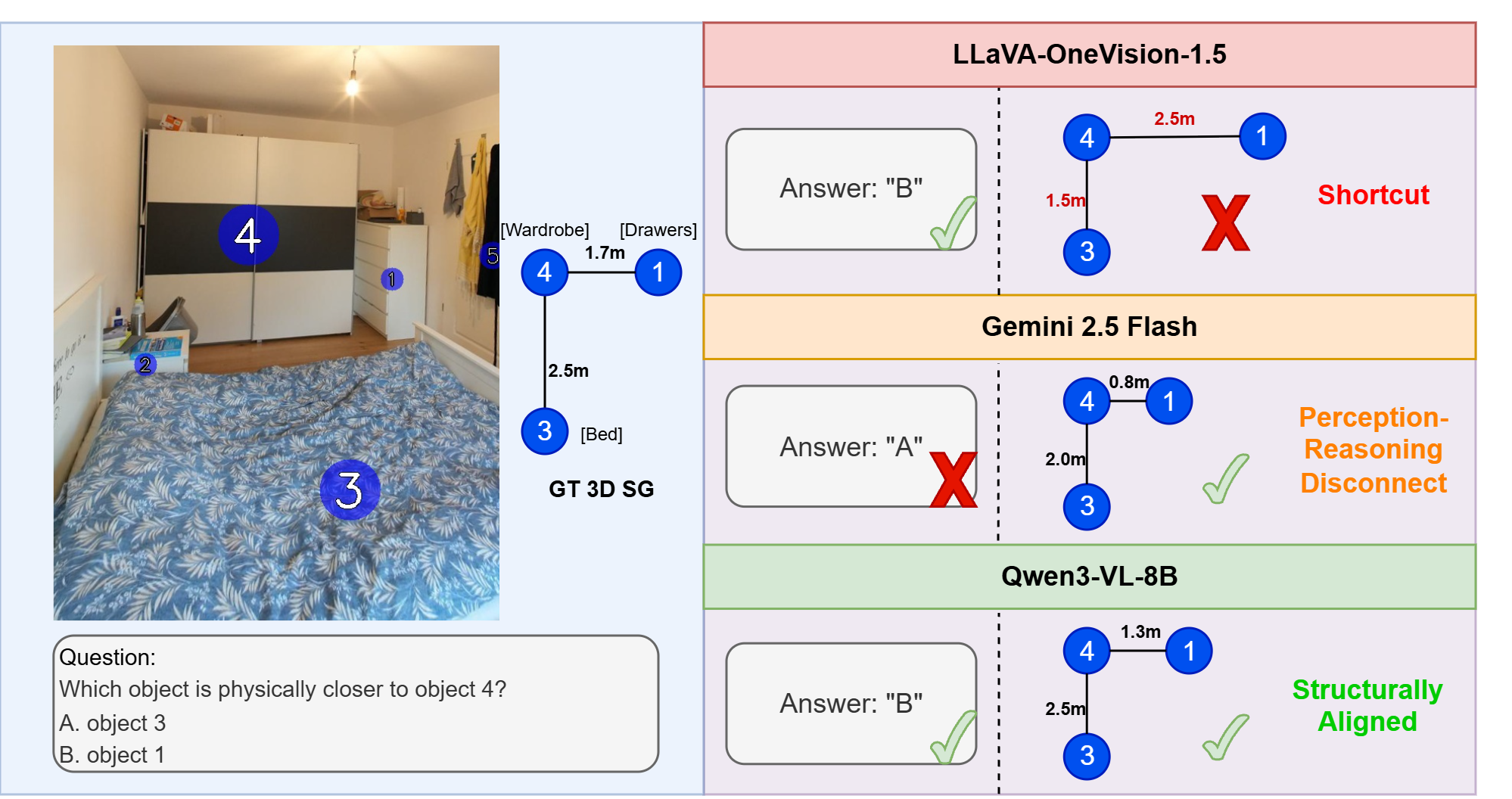}
    \caption{\textbf{Decoupling Perception from Reasoning.} Diagnosing shortcut learning and perception-reasoning disconnect.}
    \label{fig:case_study}
\end{figure}

As visualized in \cref{fig:case_study}, we ground the statistical anomalies from \cref{sec:main_results} in concrete examples. LLaVA-OneVision-1.5 visually exemplifies the \textit{Semantic Shortcut}: while its QA answer is correct, its collapsed distance estimation in the SGC confirms a reliance on linguistic priors (e.g., ``drawers are near wardrobes'') over actual geometry. Conversely, Gemini 2.5 Flash embodies the \textit{Perception-Reasoning Disconnect}: it successfully constructs a metrically plausible graph but ignores this internal structure during QA. Finally, Qwen3-VL-8B achieves \textit{Structural Alignment} with consistent QA and topology. Yet, its imperfect absolute distances indicates that while topological grounding is emerging, precise metric estimation remains a frontier for current VLMs.

This case study underscores the necessity of our generative structural probe. Traditional interpretability methods, such as Attention Maps \cite{zagoruyko2016paying} or Blind Baselines, can only determine if a model attends to an image region. In our case, an attention map would correctly highlight the drawers for LLaVA-OneVision-1.5, masking its fundamental distance blindness. By forcing the explicit readout of metric scale and topology via SGC, CRISP moves beyond checking for image usage to evaluating structural understanding, providing the granular diagnosis required for embodied applications.

\section{Unlocking Spatial Intelligence via Structural Intervention}
\begin{figure}[tb]
    \centering
    \includegraphics[width=0.75\linewidth]{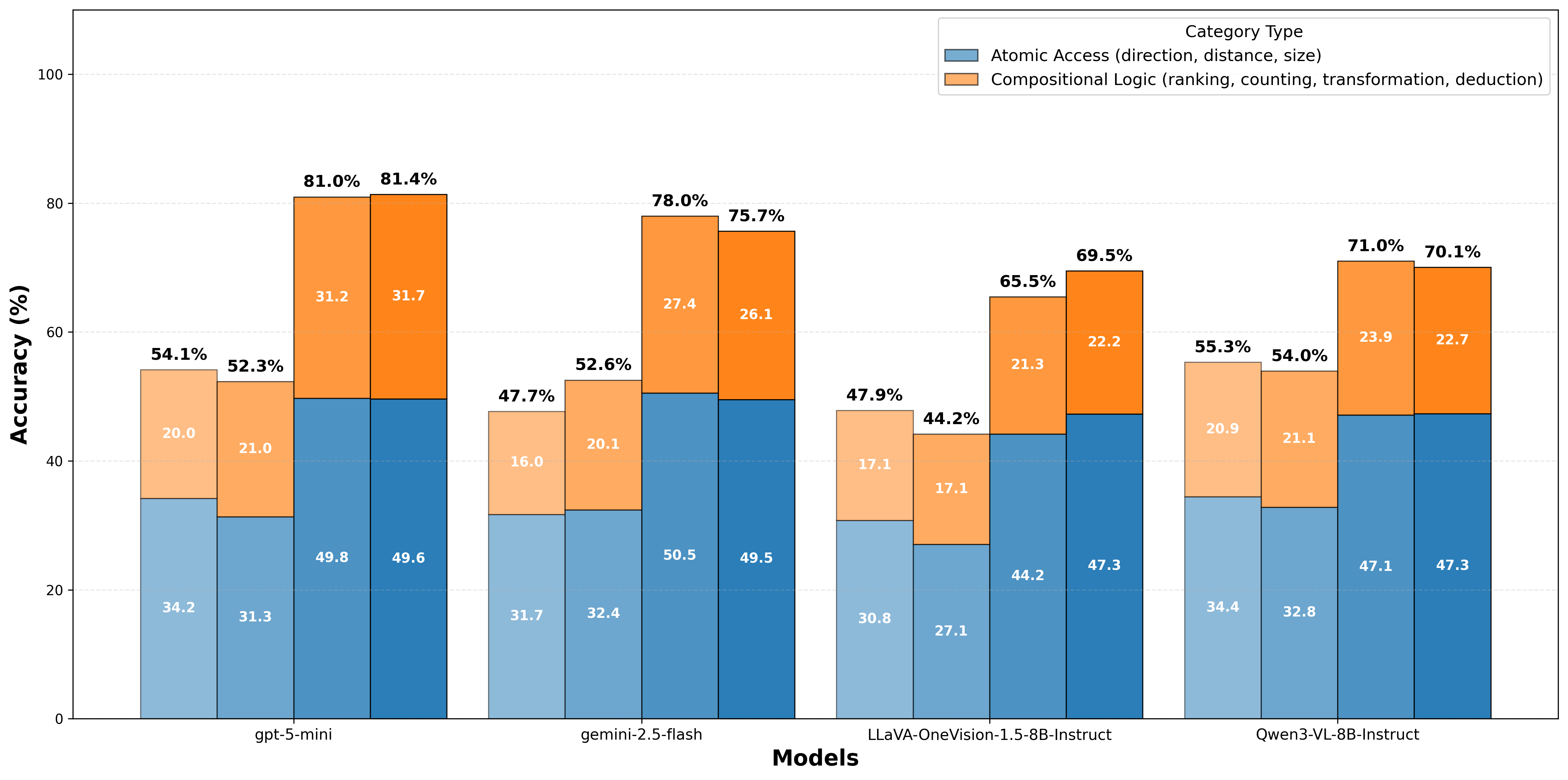}
    \caption{\textbf{Unlocking Reasoning Potential.} We compare QA accuracy across four input settings (from left to right): \textbf{Base}, \textbf{Pred SG}, \textbf{Multimodal GT SG}, and \textbf{Text-Only GT SG}. The dramatic performance surge in the two rightmost GT bars confirms the presence of robust latent reasoning engines, while the stagnation in Pred SG visually quantifies the severity of modality conflict.}
    \label{fig:stacked_bar}
\end{figure}

Our pilot ablation (Appendix C.2) demonstrates that scaling textual reasoning (e.g., CoT) primarily amplifies semantic shortcuts rather than resolving the metric grounding bottleneck. This implies that the core deficit lies not in reasoning depth, but in flawed structural perception. To validate this perception-reasoning consistency hypothesis (\cref{sec:metrics}), we must bypass the noisy visual perception system altogether. Thus, we design a dual-intervention experiment. First, by providing the Ground-Truth (GT) 3D SG alongside the image, we simulate ``idealized consistency,'' allowing the model to reason directly over perfect structured semantic representations. Conversely, feeding the model's Predicted (Pred) SGs probes its robustness against modality conflict \cite{zhang2025robust}. This comparison decisively confirms whether the VLM bottleneck is rooted in perceptual failures or inherent reasoning deficits.

\subsection{The Ceiling Analysis: Unlocking Reasoning Potential}
As illustrated in \cref{fig:stacked_bar}, injecting GT 3D SGs triggers substantial performance jumps across all models, most notably a 30.3\% surge for Gemini 2.5 Flash. To contextualize this ``oracle gain,'' we first establish a theoretical ceiling using a text-only control setting. The high accuracy ($\sim$80\%) achieved in this purely symbolic mode validates the completeness of our 3D SG formulation, confirming it encapsulates all necessary spatial information to resolve complex queries. Crucially, by injecting this ground-truth structure back into the multimodal setting, we simultaneously resolve the compounded bottlenecks diagnosed in \cref{sec:main_results}: it supplies the missing metric precision and provides an explicit structural scaffold that actively grounds the reasoning process. The notable performance recovery towards the theoretical ceiling explicitly provides compelling evidence about our core thesis: robust latent reasoning engines already exist within current VLMs; they are merely underutilized, bottlenecked by the perception-reasoning disconnect rather than intrinsic logical deficiencies.

\subsection{The Fragility of Reasoning: Modality Conflict}
\label{sec:modality_conflict}
While GT SGs unlock reasoning potential, feeding the model's own Pred SGs triggers a widespread performance collapse (\cref{fig:stacked_bar}). Far from introducing artificial noise, we posit this setting acts as a critical proxy for self-correction in long-horizon reasoning. Analogous to the intermediate reasoning steps that drive LLM planning, Pred SGs act as the explicit externalization of the VLM's latent spatial modeling process. Consequently, the observed collapse exposes a persistent trust bias, prioritizing their own erroneous linguistic externalizations over raw visual evidence. This confirms that current reasoning engines lack the robustness to resolve modality conflict, allowing generated hallucinations to override correct perceptual signals.

Gemini 2.5 Flash is the sole outlier that improves with Pred SG. This success validates a threshold hypothesis: its Pred SGs surpass a critical quality threshold (minimizing the conflict source), and its reasoning engine demonstrates superior conflict resolution, effectively using the image to verify and refine imperfect structural cues.

\subsection{Complexity Breakdown}

To determine whether structured inputs merely facilitate ``table lookup'' or genuinely empower reasoning, we categorize tasks into \textit{Atomic Access} (attribute retrieval) and \textit{Compositional Logic} (multi-hop reasoning). Injecting GT SGs reveals a pronounced cognitive divergence. Proprietary models demonstrate significant gains specifically in compositional logic ($>50\%$). This strongly indicates that their latent reasoning engines are highly capable of complex symbolic manipulation, corroborating our premise. Conversely, open-source models primarily improve on atomic tasks but stagnate in compositional logic. This isolates a fundamental deficiency in their reasoning depth: while they can parse explicit structural syntax, they lack the cognitive capacity to perform multi-hop mental simulations over these structures. 

Analyzing Pred SG inputs reveals an intriguing trade-off. Imperfect SGs degrade atomic access, corroborating the trust bias discussed in \cref{sec:modality_conflict}. However, for compositional logic, most models (except LLaVA-OneVision-1.5) actually benefit from imperfect SGs. This suggests that explicit structure, even when metrically noisy, acts as a \textit{Cognitive Scaffold.} By offloading the burden of maintaining spatial topology, it frees up computation for high-level reasoning, proving that structure itself is a key enabler for spatial intelligence.

\section{Discussion and Conclusion}
We propose CRISP as a novel structural-diagnostic paradigm that provides a conceptual roadmap for future VLM development, hypothesizing potential resource allocations based on diagnostic signatures (see Appendix D).

\begin{enumerate}
    \item \textbf{Semantic Shortcut (High QA + Low SGC/Consistency):} Models bypass geometry and rely on language priors. To pinpoint this, comparing Base vs. Multimodal GT SG input can verify if the logical engine is robust but starved of accurate perception. Consequently, resolving this bottleneck may require scaling the vision encoder to refine raw metric perception.
    \item \textbf{Perception-Reasoning Disconnect (Moderate/High SGC + Low QA):} The vision encoder extracts coarse structures, but the reasoning engine fails to anchor to them. Overcoming this failure mode likely points toward a fundamental redesign of the multimodal alignment strategy to effectively bridge this internal utilization gap.
    \item \textbf{Modality Trust Bias:} Diagnosed when QA performance drops with Predicted SG compared to the base setting. This measures the model's vulnerability to modality conflict, prioritizing self-generated text over raw visual evidence. Mitigating this bias suggests a need for upgrading the LLM backbone to deepen its compositional reasoning and conflict resolution capabilities.
\end{enumerate}
Ultimately, models must demonstrate high consistency across all modalities to validate true structural alignment, ensuring logical deductions are strictly grounded in explicitly perceived 3D geometry.

\noindent\textbf{Limitations.} First, terms like \textit{Perception-Reasoning Disconnect} denote empirical behavioral syndromes rather than proven causal mechanisms. Second, while comprehensive spatial intelligence requires dynamic multi-view reasoning (e.g., parallax), CRISP establishes static 3D grounding as its unavoidable prerequisite, leaving interactive spatiotemporal exploration for future iterations.

\noindent\textbf{Conclusion.} In this work, we introduced CRISP to evaluate visual spatial intelligence through the lens of consistency. Our evaluations reveal that traditional QA metrics often create an ``illusion of spatial intelligence'' driven by semantic shortcuts. By introducing rigorous SGC tasks and consistency analysis, we uncover a systematic \textit{Perception-Reasoning Disconnect}. Proprietary models possess robust latent reasoning engines that remain bottlenecked by coarse perceptual precision and critical alignment failures, while open-source models still struggle with intrinsic reasoning depth. 
These findings suggest that scaling end-to-end training alone is insufficient, progress requires explicitly enforcing structural consistency. By jointly evaluating QA accuracy, structural fidelity, and consistency, CRISP provides the diagnostic tools to identify whether a model genuinely grounds its spatial reasoning or merely exploits linguistic shortcuts.

\clearpage  

\section*{Acknowledgements}
The computations and data handling were enabled by resources provided by the National Academic Infrastructure for Supercomputing in Sweden (NAISS), partially funded by the Swedish Research Council through grant agreement no. 2022-06725.

%
%
\bibliographystyle{splncs04}
\bibliography{main}

\clearpage


\section*{Supplementary material}

# Supplementary Material for: From Hallucination to Grounding: Diagnosing Visual Spatial Intelligence via CRISP

Zhixing Li and Yinan Yu

Chalmers University of Technology, Gothenburg, Sweden
{zhixingl, yinan}@chalmers.se

## Contents



## A Benchmark Construction Pipeline

### A.1 Data Source and Annotations

We construct our diagnostic dataset utilizing two large-scale, real-world datasets equipped with high-quality 3D sensor annotations: nuScenes [4] (outdoor) and ScanNet++ [15] (indoor). To isolate spatial reasoning from temporal dynamics, we extract static key frames from these sequences. Initially, we randomly sample a subset of scenes from their respective training splits for further processing.

For nuScenes, we utilize images captured across all six vehicular cameras at a resolution of 1600 x 900. For ScanNet++, we employ undistorted DSLR images at a resolution of 1758 x 1152. During the 3D annotation extraction, we strictly retain semantically meaningful objects while filtering out background structural elements (e.g., walls, floors) to focus on actionable entities.

Given that VLMs lack prior knowledge of absolute spatial reference frames, we transform the original 3D bounding box coordinates of all objects from the world coordinate system to the ego-centric camera coordinate system. Crucially, with the exception of up-down spatial relationship evaluations (which must be conducted within the world coordinate system as they are strictly defined by the axis of gravity), all subsequent pipeline operations, including visibility filtering, relational graph construction, and QA generation, are executed entirely within this camera-centric coordinate space.

### A.2 3D to 2D Projection

To bridge the 3D annotations and 2D visual observations, this step is primarily executed by utilizing the official development kits provided with the respective datasets. The core objective is to derive the accurate 2D bounding boxes for each object within a given image. Acquiring these 2D boundaries enables us to ascertain an object's exact presence within the camera's field of view and calculate its specific 2D pixel dimensions, thereby establishing the essential prerequisites for the subsequent visibility and scale filtering protocols.

### A.3 Size and Visibility Filter

The primary objective of this phase is to eliminate objects that fall outside the field of view, suffer from severe truncation or occlusion, or lack sufficient pixel resolution for reliable identification. By implementing stringent, dataset-specific filtering protocols, we retain only high-quality, unambiguously visible targets, thereby minimizing confounding errors originating from the basic visual perception stage.

**NuScenes Filtering Protocol:**

1. *Resolution Filter:* We drop objects with extreme aspect ratios or diminutive scales by ensuring the shortest side of the 2D bounding box exceeds 40 pixels.
2. *Visibility Level:* We strictly utilize the official NuScenes visibility annotations, retaining only objects with visibility_level == 4 (ranging from 80% to 100% visible).
3. *Frame Clipping:* Objects whose bounding boxes fall entirely outside the image boundaries are discarded.

**ScanNet++ Filtering Protocol:** For indoor scenes, which feature dense clutter and complex occlusions, we leverage the underlying 3D mesh annotations for granular filtering.

1. *Depth Constraints:* We restrict objects to a camera-coordinate depth ($z$) range of $0.2 < z \leq 5.0$ meters. This strictly prevents severe camera-plane clipping distortions from objects positioned too close to the sensor, while discarding distant targets that lack sufficient visual detail for VLM recognition.
2. *Scale and Aspect Ratio Limits:* We filter out diminutive or abnormally shaped bounding boxes. An object's bounding box must occupy at least 0.5% of the total image area. Additionally, its aspect ratio ($w/h$) must fall within the range of $[0.2, 5.0]$ to eliminate extremely thin object fragments.
3. *Visibility and Occlusion:* We enforce robust pixel-level and topological visibility. At least 5% of the object's 3D mesh vertices must be physically visible in the camera view.
4. *Border Truncation Penalty:* To prevent severe information loss from boundary cropping, if a bounding box touches the image border (within a 5-pixel margin), we mandate a strict object completion rate of $\geq 0.6$. Objects completely within the frame are not subject to this specific penalty.

For both datasets, any image frame that retains zero valid objects after executing the object-level filtering sequence is entirely discarded from the benchmark. To prepare the surviving images for the VLM evaluation, we apply a specific visual prompting strategy. Rather than using standard bounding boxes, a semi-transparent numerical ID enclosed in a minimal circular marker is rendered precisely at the center of each retained object's 2D bounding box.

This specific design choice is driven by rigorous diagnostic motivations. Primarily, it bypasses the object detection bottleneck, ensuring that any downstream failures are strictly attributed to deficits in spatial reasoning rather than mere perception recall errors. Furthermore, this numerical referencing prevents semantic information leakage. By abstracting explicit class names into neutral IDs, we force the VLMs to physically ground the specific target in the scene rather than relying on linguistic co-occurrence priors. Most crucially, intentionally eschewing full 2D bounding boxes in favor of localized center markers eliminates 2D scale leakage. If full bounding boxes were provided, models could trivially estimate an object's relative distance or physical size based solely on its 2D pixel area. Our point-based markers effectively neutralize this shortcut, thereby enforcing genuine 3D spatial reasoning.

### A.4 Master 3D Scene Graph Construction and Scene Balancing

In this stage, we utilize the images refined during the preceding visibility filtration phase to construct the Master 3D Scene Graphs and strictly balance the dataset to ensure representational diversity. The pipeline is executed in three sequential phases:

**1. Node-Level Complexity Balancing.** To guarantee a diverse distribution of scene complexities and prevent evaluation bias toward trivial scenarios, we categorize images into three distinct bins based on their valid object count: Sparse (2-3 objects), Moderate (4-8 objects), and Dense (9-12 objects).

We enforce a strict dataset-level distribution ratio of approximately $1:2:2$ across these bins. Consequently, exactly 600 highly-curated images satisfying this complexity distribution are preserved from both the nuScenes and ScanNet++ datasets, establishing a balanced foundation of 1,200 scenes.

**2. Topological Construction and Relational Formalization.** For each retained scene, we formulate the explicit nodes and edges of the Master 3D SG.

- *Node Formulation:* We extract the semantic category, 3D dimensions, egocentric camera coordinates, and gravity-aligned world coordinates directly from the annotations.
- *Topology Generation:* To establish a cohesive local topology, we compute the spatial centroid of all valid objects within the scene. The object nearest to this centroid is designated as the *central anchor*. We then compute the 3D Euclidean distances from this anchor to all other nodes, establishing directed edges exclusively to targets within a 25-meter radius. This constraint effectively precludes the formation of semantically meaningless, ultra-long-range relationships.
- *Relational Formalization:* We assign strict deterministic semantic predicates to each edge based on relative spatial geometry. To prevent borderline ambiguity, a margin $\epsilon$ is applied. Let the anchor be $A$ and the target be $B$.
  - Horizontal Relations (Ego-Centric): Evaluated using the camera coordinate system (where $+X$ is right, $+Z$ is forward). Given the centroid coordinate differences $\Delta x = x_B - x_A$ and $\Delta z_{\text{cam}} = z_B - z_A$:

$$\texttt{Right}: \Delta x > \epsilon \qquad \texttt{Left}: \Delta x < -\epsilon$$
$$\texttt{Behind}: \Delta z_{\text{cam}} > \epsilon \qquad \texttt{In Front Of}: \Delta z_{\text{cam}} < -\epsilon$$

  - Vertical Relations (Gravity-Aligned): Evaluated using the absolute World $Z$ coordinates. To account for varying object heights and physical stacking, we utilize the absolute top ($Z_{\text{max}}$) and bottom ($Z_{\text{min}}$) bounds of the 3D bounding boxes rather than centroids:

$$\texttt{Up}: Z^B_{\text{min}} - Z^A_{\text{max}} > -\epsilon \quad \text{(Target's bottom is higher than Anchor's top)}$$
$$\texttt{Down}: Z^B_{\text{max}} - Z^A_{\text{min}} < \epsilon \quad \text{(Target's top is lower than Anchor's bottom)}$$

**3. Edge-Level Redundancy Filtration.** Upon completing the edge construction, we prune any edges lacking valid spatial predicates and subsequently discard any images with empty edge sets. Finally, to eliminate semantic redundancy, we implement an edge-based diversity filter. We define a holistic edge signature for each image formatted as a set of tuples: `(from_category, to_category, relation_vector, distance_bin)`. We compute the Jaccard similarity between the edge signature sets of different images. If the similarity coefficient between any pair exceeds 0.8, we isolate the redundancy by retaining solely the image containing the greater number of edges, ensuring maximal relational diversity.

Upon completion of the comprehensive Master 3D SG construction and edge-level filtration, the pipeline yields a final curated set of 1,162 high-quality images.

The minor attrition from the initial 1,200 balanced scenes is a direct result of our strict filters. Notably, we deliberately abstained from implementing global visual similarity filtering (e.g., CLIP-based deduplication); consequently, the dataset naturally encompasses multiple distinct camera viewpoints of identical physical environments.

We argue this is a crucial diagnostic feature rather than a limitation. While these multi-view scenes share high visual homology, their ego-centric topological structures (the Master 3D SGs) exhibit substantial variance due to perspective shifts. From a diagnostic standpoint, evaluating a model on visually similar scenes with divergent structured representations offers significantly greater discriminatory power than testing on visually disparate scenes that share identical spatial layouts. This paradigm functions as a rigorous scene-level hard negative test. It actively penalizes VLMs that rely on 2D visual memorization or global scene semantics, explicitly verifying their capacity to dynamically ground nuanced, viewpoint-dependent spatial transformations in the 3D physical world.

### A.5 Spatial QA and SGC Templates

In this section, we present the prompt templates employed throughout the evaluation process, as shown in Tab. 1. Notably, within the spatial QA tasks, the MCQs encompass both binary and four-option formats.

### A.6 QA Generation and Logic Engine

To prevent VLMs from exploiting dataset artifacts or statistical shortcuts, our procedural generation pipeline enforces three core mechanisms.

1. **Semantic Constraint Filtration.** For MCQs, we avoid naive random sampling of distractors to prevent shortcut learning. For example, if the ground-truth relation is a composite direction (e.g., `front-left`), the engine strictly forbids overlapping atomic distractors (e.g., `front` or `left`), while specifically permitting orthogonal composites (e.g., `back-right`). This rigorous negative sampling forces the model to perform precise multi-axis geometric reasoning, ensuring it cannot solve queries through superficial keyword matching or simple elimination.
2. **Granularity-Aware Metric Margins.** To eliminate borderline ambiguity, we apply property-specific metric rules. For comparative distance queries, spatial separation relies on 3D centroids. Since centroid distance inherently conflates physical separation with object volume, we enforce a strict $\Delta d >$ 1.0 meter margin to guarantee unambiguous depth ordering. Conversely, intrinsic size comparisons compute exact mathematical differences directly from high-fidelity 3D bounding boxes without margins. This intentionally challenges models to overcome 2D perspective distortions and demonstrate fine-grained sensitivity to absolute 3D scale.
3. **Quantitative Statistical Balancing.** To neutralize VLM positional biases and linguistic priors, our engine executes a rigorous global balancing pass

across the benchmark. The final dataset comprises an optimized split of 73.7% MCQs and 26.3% NAQs. We achieve strictly balanced answer distributions: binary MCQs reflect a 46.9% to 53.1% split, and quad-choice MCQs exhibit near-perfect uniformity across all option slots (A: 24.3%, B: 25.3%, C: 25.8%, D: 24.6%). This mathematically verified uniformity guarantees that models cannot artificially inflate their scores via statistical guessing.

**Table 1: Spatial QA Prompt Templates.**

| **Category** | **Type** | **Example template** |
|---|---|---|
| Directional Perception | MCQ | "From the camera's viewpoint, is <OBJ_A> in front of <OBJ_B>?"<br>"From the camera's viewpoint, which best describes the relative position of <OBJ_A> to <OBJ_B>?" |
| Size Estimation | MCQ | "Considering real-world physical dimensions, which object is larger in volume: <OBJ_A> or <OBJ_B>?"<br>"Which object is taller: <OBJ_A> or <OBJ_B>?" |
| | NAQ | "Approximately how long (in meters) is <OBJ_A>?" |
| Distance Estimation | MCQ | "Which object is physically closer to the camera: <OBJ_A> or <OBJ_B>?"<br>"Which object is physically closer to <OBJ_A>: <OBJ_B> or <OBJ_C>?" |
| | NAQ | "What is the distance (in meters) between <OBJ_A> and <OBJ_B>?" |
| Counting | MCQ | "From the camera's viewpoint and only considering the marked objects, how many objects are <REL_PHRASE> <OBJ_A>?"<br>"Considering the marked objects, how many objects are within <DIST> meters of <OBJ_A>?" |
| Spatial Ranking | MCQ | "Rank the following objects by distance from the camera (nearest to farthest): <OBJ_A>, <OBJ_B>, <OBJ_C>."<br>"Rank the following objects by height (tallest to shortest): <OBJ_A>, <OBJ_B>, <OBJ_C>." |
| Logical Deduction | NAQ | "Suppose we place a new object A in the <DIR> direction from <OBJ_A> at a distance of approximately <DIST> meters. What is the horizontal distance (in meters) between A and <OBJ_B>?" |
| View Transformation | MCQ | "Imagine you are standing at the location of <OBJ_A> and facing directly towards <OBJ_B>. From this perspective, in which direction is <OBJ_C> located relative to you?" |

**Prompt for 3D Scene Graph Construction**

You are given a real world image where each object is marked with a numeric id (Object 1, Object 2, ...).
**Task:**
Generate a 3D scene graph JSON that describes the spatial layout of the scene.
**Requirements:**
1. Create exactly one entry in “objects” for **every** id listed in “Scene Objects”.
2. Create edges from the **Center Object** to **all other** objects.
3. Analyze the spatial relationship from the center object to each target object.
4. For “relation”, list **ALL** applicable terms (e.g., [“left”, “in front of”]).
5. For “relation”, use ONLY these words: [“left”, “right”, “in front of”, “behind”, “up”, “down”].
**Output Format:**

```
```json{
  "objects": [
    {
      "id": int,
      "dist_to_cam": float,
      "size": {
        "w": float,
        "l": float,
        "h": float
      }
    }
  ],
  "edges": [
    {
      "from": int,
      "to": int,
      "distance": float,
      "relation": ["string"]
    }
  ]
}```
```

**Input Data:**
Center Object id: `{center_id}`
Scene Objects (ids shown in the image): `{object_list_str}`
**Response:**
Output the valid JSON only, without any extra text.
```

### A.7 Validation of the SG Solver and Theoretical Limits

To rigorously validate our consistency metric, we formalize the derivation process of our rule-based solver, which operates across three distinct tiers of cognitive complexity:

**1. Atomic Access.** For foundational queries concerning direct perception (e.g., Size, Distance, and absolute Direction), the solver performs a direct $O(1)$ attribute retrieval and threshold-based matching on the parsed graph nodes and edges.

**2. Basic Compositional Logic.** For structural queries such as Counting and Spatial Ranking, the derivation strictly mirrors the execution of our deterministic QA generation engine. The solver executes algorithmic sorting or counting over the retrieved graph topology and performs inverse matching against the multiple-choice options.

**3. Mental Simulation.** For advanced Compositional Logic queries requiring continuous spatial imagination (e.g., Logical Deduction and View Transformation), the solver mathematically reconstructs a local geometric space in the $xz$-plane to derive answers.

1. *Logical Deduction:* Let the reference object $A$ be the origin $(0, 0)$. For an existing object $B$, the solver parses its relative direction set into a unit vector $\boldsymbol{u}_{AB}$ and retrieves its distance $d_{AB}$, establishing its coordinate as $P_B = d_{AB}\boldsymbol{u}_{AB}$. Given a hypothetical object $H$ placed at distance $d_H$ along direction $\boldsymbol{u}_H$ relative to $A$, the solver analytically deduces the unseen distance between $B$ and $H$ via the $L_2$ norm:

$$d_{BH} = \|d_H\boldsymbol{u}_H - d_{AB}\boldsymbol{u}_{AB}\|_2$$

2. *View Transformation:* Given unit direction vectors $\boldsymbol{v}_{AB}$ and $\boldsymbol{v}_{AC}$ mapped from the graph's edge predicates, the solver computes their continuous azimuthal angles: $\theta_B = \text{atan2}(\boldsymbol{v}^x_{AB}, \boldsymbol{v}^z_{AB})$ and $\theta_C = \text{atan2}(\boldsymbol{v}^x_{AC}, \boldsymbol{v}^z_{AC})$. The continuous relative rotation $\Delta\theta = (\theta_C - \theta_B) \pmod{2\pi}$ is mapped into a discrete geometric sector, which is then translated back to canonical semantic predicates.

Because the solver evaluates model-generated imperfect 3D SGs, it is designed with strict fallback mechanisms. If the parsed graph is missing necessary nodes or edge relations required for a query, or if the mathematically derived relation fails to match any provided QA option, the solver explicitly halts and outputs a `[FAILED]` token. This mechanism ensures that structurally incomplete, hallucinated, or ungrounded graphs are directly penalized during the consistency evaluation.

Empirically, the formalized derivation protocol described above successfully resolves approximately 95% of the benchmark queries. Crucially, this 5% systemic margin is exclusively localized to advanced "mental simulation" tasks (View Transformation 2.46%, Logical Deduction 2.35%). For all foundational atomic access and basic compositional logic, the solver achieves perfect derivation on Ground Truth 3D SGs, yielding answers strictly identical to those of

the original QA generator. Therefore, the remaining 5% error is not a failure of the representation, but rather exposes the inherent boundary of applying rigid mathematical rules to a lightweight symbolic abstraction. By design, 3D SGs intentionally compress continuous visual environments into discrete topological and metric nodes. While this abstraction is highly efficient, it inevitably discards fine-grained spatial continuity. Consequently, in some cases during mental simulation, such as dense object clustering, the compressed SG lacks the nuanced geometry required for flawless resolution. While VLMs can utilize raw pixel-level cues to intuitively navigate these crowded regions, our rule-based solver performs strictly "blind" derivations. Operating entirely without visual aids, it relies exclusively on the compressed symbolic text, and thus cannot perfectly simulate these highly dense physical interactions. We deliberately choose *not* to filter out this 5% subset from our evaluation, driven by three rigorous methodological principles:

1. **Preserving the Difficulty Ceiling.** The 5% of instances that elude simple rule-based derivation represent the most challenging scenarios in spatial reasoning, requiring fine-grained continuous physical intuition. Artificially removing these instances simply because our hand-crafted logic engine cannot perfectly solve them would improperly lower the benchmark's difficulty ceiling. We retain these edge cases to rigorously test the VLMs' end-to-end capability limits when visual inputs are present.

2. **The Synergy of Symbolic and Continuous Representations.** Rather than viewing this 5% margin as a flaw, we treat it as a quantitative measure of the semantic gap between explicit symbolic execution and implicit visual reasoning. 3D SGs provide the indispensable structural scaffold for spatial intelligence. However, solving the most extreme spatial edge cases requires synergizing this explicit structural knowledge with un-externalizable, continuous visual details. Retaining these queries acknowledges that while explicit grounding (Consistency) is crucial, perfect spatial intelligence ultimately requires multi-modal feature fusion beyond simple rigid rules.

3. **Metric Robustness and Uniform Systemic Dampening.** A critical concern is whether the solver's imperfection on these complex queries inherently invalidates the Consistency Score. It is true that for this specific 5% subset, a VLM might correctly answer the QA relying on continuous visual features, yet receive an "inconsistent" penalty because the solver cannot derive the answer from the discrete SG. However, statistically, this acts as a uniform systemic dampener across all evaluated models. Because this noise margin is small ($< 5\%$) and uniformly distributed, it serves as a strict lower bound for explicit consistency without altering the relative performance rankings or the macroscopic diagnostic conclusions of our benchmark.

**Table 2:** Detailed evaluation results of Spatial QA task. Bold denotes the best performance within each category.

| | MCQ | | | | | | NAQ | | | |
|---|---|---|---|---|---|---|---|---|---|---|
| **Model** | **Counting** | **Direction** | **Distance** | **Ranking** | **Size** | **Transformation** | **Deduction** | **Distance** | **Size** | **Overall** |
| ***Proprietary Models*** | | | | | | | | | | |
| Gemini-2.5-Flash [5] | 53.63 | 64.82 | 73.86 | 30.67 | 67.38 | 17.38 | 41.99 | 32.88 | 46.49 | 47.68 |
| Gemini-2.5-Pro † [5] | **65.65** | **77.54** | **85.12** | **59.01** | **70.14** | 32.44 | **46.63** | 43.35 | **50.46** | **58.93** |
| Gemini-3-Flash [6] | 62.36 | 72.46 | 79.75 | 43.86 | 56.88 | **34.94** | 38.76 | 43.54 | 48.14 | 53.41 |
| GPT-5-Mini† [9] | 63.72 | 75.93 | 79.55 | 50.67 | 66.52 | 22.46 | 42.78 | 38.69 | 47.01 | 54.15 |
| GPT-5.2 [9] | 45.35 | 71.14 | 74.90 | 29.96 | 61.79 | 24.58 | 30.53 | **51.83** | 48.71 | 48.75 |
| ***Open-source Models*** | | | | | | | | | | |
| LLaVA-OV-1.5-8B [1] | 49.43 | 71.04 | 75.62 | 41.61 | 60.07 | 20.70 | 41.82 | 32.69 | 37.70 | 47.85 |
| InternVL3.5-8B [10] | 50.91 | 69.57 | 74.79 | 43.50 | 62.91 | 28.28 | 47.30 | 39.06 | 18.97 | 48.37 |
| InternVL3.5-38B [10] | **52.38** | **73.19** | 83.99 | 46.82 | 66.44 | 23.20 | 48.52 | 47.28 | **43.99** | 53.98 |
| Qwen2.5-VL-7B [3] | 47.51 | 66.78 | 73.45 | 36.32 | 61.27 | 20.98 | 38.63 | 32.22 | 25.85 | 44.78 |
| Qwen3-VL-8B [2] | 47.73 | 71.33 | 85.95 | **52.65** | 65.49 | 35.21 | **52.50** | 44.96 | 42.23 | **55.34** |
| Qwen3-VL-32B [2] | 45.35 | 72.65 | **87.81** | 39.82 | **67.99** | **35.67** | 47.29 | **47.70** | 41.55 | 53.98 |
| ***Specialized Baselines*** | | | | | | | | | | |
| Cambrian-S [13] | **51.47** | 55.48 | 60.23 | 32.91 | **62.39** | **38.45** | **36.28** | **41.03** | **35.32** | **45.95** |
| VG LLM [16] | 41.84 | **59.78** | **78.41** | **37.49** | 57.49 | 19.13 | 22.23 | 16.93 | 11.72 | 38.34 |

†: Thinking mode restricted.

# B Detailed Experimental Results

## B.1 Detailed Spatial QA Results

The fine-grained results of the Spatial QA evaluation across all cognitive categories are presented in Tab. 2. Analyzing this category-specific performance reveals several critical diagnostic insights that empirically substantiate the core claims made in the main text.

First, state-of-the-art VLMs already exhibit strong capabilities in basic relative spatial judgments. As demonstrated, leading models such as Gemini 2.5 Pro and Qwen3-VL-32B achieve highly commendable accuracy in relative MCQ tasks, scoring over 85% in Distance and over 70% in Direction and Size comparisons. However, a significant metric gap emerges when transitioning from relative comparisons to absolute numerical estimation within the exact same categories. For example, Qwen3-VL-32B's accuracy plummets from 87.81% on Distance MCQs to merely 47.70% on Distance NAQs. This severe degradation empirically validates that current VLMs lack precise metric grounding, relying heavily on qualitative visual priors rather than rigorous 3D scale estimation.

Furthermore, severe bottlenecks persist in complex reasoning categories that demand continuous spatial simulations and multi-hop topological inferences. Performance on View Transformation is notably deficient across the board, with even the most advanced proprietary models (e.g., Gemini 2.5 Pro at 32.44%) failing to exceed the 35% threshold. This systemic failure corroborates our core assertion: while models can parse explicit spatial linguistics, they currently lack the implicit capacity for reliable mental rotation and complex logical deduction in 3D physical spaces.

## B.2 Detailed SGC Results

The comprehensive evaluation results for the 3D SGC task are presented in Tab. 3. Analyzing these sub-metrics reveals a consistent perceptual bias across the evaluated models: they demonstrate significantly higher proficiency in qualitative semantic relationship prediction compared to absolute numerical estimation. For instance, Gemini 3 Flash achieves a robust 72.33% in Relationship (Rel) accuracy, yet scores only 57.80% in Camera Distance (Cam Dist) estimation. This discrepancy directly corroborates our Spatial QA finding.

A pivotal metric presented in this table is the 3D SG Format Compliance Rate. Under our exceptionally stringent parsing criteria, all proprietary models achieved 100% compliance rate, and the latest open-source foundation models consistently attained rates exceeding 99%. This overwhelmingly high success rate empirically underpins the conclusions drawn from Sec. 4.2. It definitively isolates the source of SGC failures: the models are fully capable of generating complex, syntactically correct JSON structures. Consequently, their low overall SGC scores stem fundamentally from a lack of visual spatial perception rather than a deficiency in instruction following.

**Table 3:** Detailed evaluation results of SGC. The definitions for all evaluated metrics are detailed in Sec. 3.3. Specifically, Overall denotes the SGC Score, while Comp Rate refers to the 3D SG format compliance rate. A generation is classified as "compliant" if and only if the model outputs a syntactically correct JSON structure without omitting any nodes or edges. Bold denotes the best performance.

| Model | Size | Cam Dist | Dist | Est | Rel | Overall | Comp Rate |
|---|---|---|---|---|---|---|---|
| ***Proprietary Models*** | | | | | | | |
| Gemini-2.5-Flash | 69.17 | 58.59 | 55.97 | 61.24 | 68.61 | 64.92 | 100.00 |
| Gemini-2.5-Pro† | **69.85** | 44.14 | 59.38 | 57.79 | 71.36 | 64.58 | 100.00 |
| Gemini-3-Flash | 69.69 | 57.80 | **68.26** | 65.25 | **72.33** | **68.79** | 100.00 |
| GPT-5-Mini† | 68.25 | 59.73 | 56.20 | 61.39 | 72.01 | 66.70 | 100.00 |
| GPT-5.2 | 68.60 | **64.29** | 67.12 | **66.67** | 64.61 | 65.64 | 100.00 |
| ***Open-source Models*** | | | | | | | |
| LLaVA-OV-1.5-8B | 45.59 | 44.67 | 43.33 | 44.53 | 56.12 | 50.33 | 99.91 |
| InternVL3.5-8B | 60.65 | 46.34 | 46.98 | 51.32 | 61.10 | 56.21 | 99.91 |
| InternVL3.5-38B | **65.43** | 49.01 | 44.93 | 53.12 | 64.68 | 58.90 | 100.00 |
| Qwen2.5-VL-7B | 52.06 | 35.78 | 34.02 | 40.62 | 44.79 | 42.70 | 87.87 |
| Qwen3-VL-8B | 59.97 | 59.57 | 51.72 | 57.09 | **67.62** | 62.35 | 99.48 |
| Qwen3-VL-32B | 63.64 | **60.90** | **56.09** | **60.21** | 64.69 | **62.45** | 100.00 |
| ***Specialized Baselines*** | | | | | | | |
| Cambrian-S | **54.58** | **44.62** | **41.09** | **46.76** | 47.50 | 47.13 | 91.48 |
| VG LLM | 46.69 | 38.57 | 36.56 | 40.61 | **53.85** | **47.23** | 77.54 |

†: Thinking mode restricted.

Finally, the detailed metrics of the specialized spatial baselines (Cambrian-S and VG LLM) provide further granularity to our discussion regarding end-to-end post-training. As established in Sec. 4.1, optimizing solely for downstream QA performance fails to synthesize missing spatial modeling capabilities, often introducing volatile structural formatting behaviors at the direct expense of genuine 3D cognition.

### B.3 Detailed Consistency Results

Tab. 4 shows the comparison of Base QA and Derived QA Scores. Tab. 5 delineates the fine-grained Consistency Scores across all evaluated dimensions. As anticipated, the general trends mirror the Spatial QA performance, yet they expose a far more severe underlying vulnerability. Models maintain moderate structural alignment on fundamental comparative tasks, such as Gemini 2.5 Pro achieving 82.85% consistency on Distance MCQs. However, their consistency catastrophically collapses on tasks requiring continuous spatial simulations. For View Transformation and Logical Deduction, even the most advanced models struggle to surpass the 40% threshold. This degradation firmly substantiates the perception-reasoning disconnect highlighted in Sec. 4.1. It empirically demonstrates that when confronted with complex spatial logic, models tend to abandon their explicit structural grounding, relying instead on unverified implicit features or statistical hallucinations to formulate their text responses.

**Table 4: Comparison of Base QA and Derived QA Scores.**

| Model | Base QA | Derived QA |
|---|---|---|
| ***Proprietary Models*** | | |
| Gemini-2.5-Flash | 47.68 | 55.03 (+7.35) |
| Gemini-2.5-Pro | 58.93 | 55.61 (-3.32) |
| Gemini-3-Flash | 53.41 | 59.36 (+5.95) |
| GPT-5-Mini | 54.15 | 56.43 (+2.28) |
| GPT-5.2 | 48.75 | 58.77 (+10.02) |
| ***Open-source Models*** | | |
| LLaVA-OV-1.5-8B | 47.85 | 41.95 (-5.90) |
| InternVL3.5-8B | 48.37 | 46.40 (-1.97) |
| InternVL3.5-38B | 53.98 | 48.66 (-5.32) |
| Qwen2.5-VL-7B | 44.78 | 37.75 (-7.03) |
| Qwen3-VL-8B | 55.34 | 52.83 (-2.51) |
| Qwen3-VL-32B | 53.98 | 54.70 (+0.72) |
| ***Specialized Baselines*** | | |
| Cambrian-S | 45.95 | 37.79 (-8.16) |
| VG LLM | 38.34 | 34.90 (-3.44) |

Beyond the simple-versus-complex dichotomy, these detailed sub-metrics provide a crucial supplement to our main text by revealing a profound detachment in absolute numerical reasoning. A direct comparison between MCQ and NAQ consistency within the exact same category is highly revealing. For instance, while Qwen3-VL-32B maintains a 79.75% consistency for relative Distance MCQs, this metric plummets to 47.10% for absolute Distance NAQs; a similarly drastic drop is observed in Gemini 2.5 Pro (from 82.85% to 28.64%). This specific divergence proves that while VLMs can occasionally align their qualitative topological predictions with their structured outputs, their internal representations of absolute physical scale are entirely decoupled from their numerical estimations. When models generate exact distances or sizes, the outputs are largely disconnected from their own explicitly perceived 3D coordinates.

**Table 5:** Detailed evaluation results of Consistency Score. Bold denotes the best performance within each category.

| Model | MCQ | | | | | | NAQ | | | |
|---|---|---|---|---|---|---|---|---|---|---|
| | **Counting** | **Direction** | **Distance** | **Ranking** | **Size** | **Transformation** | **Deduction** | **Distance** | **Size** | **Overall** |
| ***Proprietary Models*** | | | | | | | | | | |
| Gemini-2.5-Flash | 43.20 | 66.78 | 74.69 | 24.75 | 73.75 | 25.32 | 38.40 | 46.16 | 56.78 | 49.98 |
| Gemini-2.5-Pro† | 58.39 | 78.23 | **82.85** | **51.66** | **79.78** | 32.26 | 43.55 | 28.64 | **60.14** | **57.28** |
| Gemini-3-Flash | **61.00** | 74.76 | 81.51 | 38.83 | 65.66 | **39.09** | 34.43 | 45.15 | 56.35 | 55.20 |
| GPT-5-Mini† | 53.40 | **80.72** | 79.44 | 42.42 | 73.92 | 28.84 | **43.57** | 52.04 | 56.35 | 56.75 |
| GPT-5.2 | 38.66 | 76.66 | 76.76 | 22.87 | 64.20 | 24.21 | 31.02 | **52.39** | 53.04 | 48.87 |
| ***Open-source Models*** | | | | | | | | | | |
| LLaVA-OV-1.5-8B | 37.19 | 60.67 | 63.74 | 28.88 | 64.63 | 14.23 | 33.23 | 22.83 | 25.43 | 38.98 |
| InternVL3.5-8B | 40.82 | 61.06 | 67.77 | 31.03 | 68.67 | 19.04 | 42.52 | 38.29 | 23.13 | 43.59 |
| InternVL3.5-38B | **41.95** | 61.89 | 68.08 | 29.15 | **73.32** | 27.08 | 49.79 | 42.11 | 44.97 | 48.70 |
| Qwen2.5-VL-7B | 32.43 | 52.94 | 60.54 | 32.02 | 48.71 | 12.38 | 30.39 | 21.08 | 30.13 | 35.62 |
| Qwen3-VL-8B | 39.91 | 63.89 | 78.72 | **38.92** | 66.01 | 26.06 | **53.15** | **56.56** | **46.22** | **52.16** |
| Qwen3-VL-32B | 39.57 | **72.90** | **79.75** | 28.88 | 70.14 | **29.94** | 50.17 | 47.10 | 45.63 | 51.56 |
| ***Specialized Baselines*** | | | | | | | | | | |
| Cambrian-S | **42.29** | 37.38 | 49.28 | 19.37 | **54.30** | **20.15** | **25.49** | 25.71 | 33.97 | 34.22 |
| VG LLM | 21.66 | **42.86** | **55.58** | **36.50** | 38.12 | 15.16 | 15.79 | **55.74** | **36.17** | **35.29** |

†: Thinking mode restricted.

### B.4 Consistency Confusion Matrix Analysis

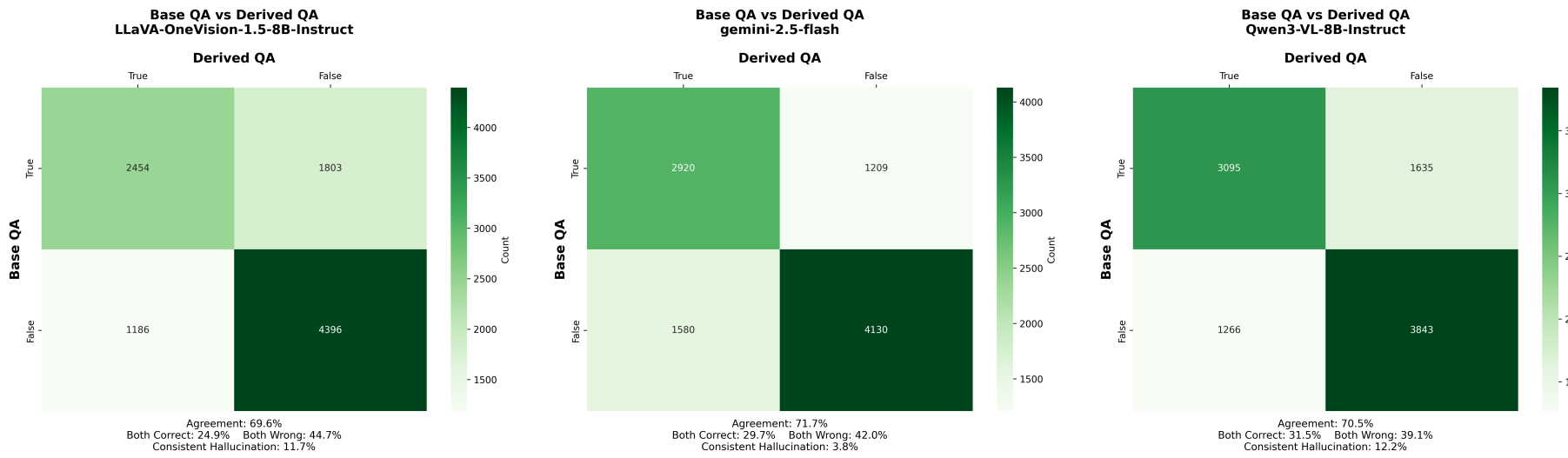


**(a)** LLaVA-OneVision-1.5-8B **(b)** Gemini-2.5-Flash **(c)** Qwen3-VL-8B

**Fig. 1:** Confusion matrices comparing Base QA and Derived QA predictions.

To empirically ground the diagnostic archetypes identified in Sec. 4.3, we visualize the consistency evaluation protocol via instance-level confusion matrices (Base QA vs. Derived QA) for three representative models. As shown in Fig. 1, the distribution of both off-diagonal and joint-failure elements visually distinguishes the fundamental failure modes:

1. **Semantic Shortcut (LLaVA-OneVision-1.5, Fig. 1a):** Exhibits a heavy concentration in the off-diagonal where Base QA is True but Derived QA is False, explicitly quantifying its reliance on linguistic priors over structural grounding.
2. **Perception-Reasoning Disconnect (Gemini 2.5 Flash, Fig. 1b):** Displays a prominent volume of instances where Derived QA is True while Base QA fails. More critically, within the joint-failure quadrant (False-False), its consistent hallucination rate is anomalously low (3.8%) despite a massive 42.0% joint-failure volume. This explicitly reveals that when the model fails, its text and spatial modules hallucinate incoherently in entirely divergent directions, proving the reasoning engine does not anchor to its internal visual representations.
3. **Emerging Structural Alignment (Qwen3-VL-8B, Fig. 1c):** Confirms our theoretical boundaries. The consistent hallucination accounts for 12.2% of the subset. Unlike Gemini's fragmented failures, Qwen's coherent errors prove that its reasoning engine is tightly coupled with its visual perception, systematically following flawed geometry to a logical conclusion. However, this alignment is not yet absolute. A non-trivial cluster in the off-diagonal (e.g., a substantial volume where Base QA is True but Derived is False) reveals persistent residual shortcut behaviors. This explicitly underscores that while frontier open-source models are transitioning towards genuine grounding, perfectly resisting the gravitational pull of language priors remains an ongoing challenge for the community.

### B.5 Extended Case Studies

As a complement to the qualitative analysis in Sec. 4.3, this section provides further instance-level case studies to empirically substantiate our core findings. Specifically, we meticulously examine three representative diagnostic scenarios: *Consistent Hallucination* (Fig. 2), *Imperfect Alignment* (Fig. 3), and *Solver Failure* (Fig. 4). To offer a transparent view of the models' internal cognitive processes, each case explicitly contrasts the VLM's end-to-end QA output with a snippet of its raw, explicitly generated 3D SG in JSON format. Accompanied by targeted diagnostic insights, this granular format explicitly traces how specific metric or topological errors in the symbolic space propagate (or fail to propagate) into the final textual reasoning.

### B.6 Extended Results for Structural Intervention Task

Tab. 6 provides a granular breakdown of model performance across different intervention settings, revealing distinct cognitive signatures that average scores often obscure. To visualize the instance-level dynamics of these interventions, we present a case study of GPT 5 Mini in Fig. 5.

**The Oracle Potential and Systematic Alignment Failures.** Under GT 3D SG intervention, proprietary models demonstrate robust latent reasoning capabilities. As shown in Tab. 6, GPT 5 Mini achieves near-perfect precision in metric-heavy tasks, reaching 100.00% in NAQ Distance estimation. This is visually corroborated by Fig. 5c, where the model's median MRA jumps from 0.400 to 1.000. However, this symbolic prowess is sharply contradicted by an anomaly in the Direction category, where GPT 5 Mini's score plummets to 48.83%, which suggests a systematic semantic misalignment rather than a mere reasoning failure. It implies that the model may be interpreting the spatial predicates under a conflicting reference frame definition (e.g., confusing camera-centric vs. object-centric coordinates) compared to the ground truth annotations. Unlike Qwen3-VL-8B (71.53%), which aligns well with the provided definitions, GPT 5 Mini appears to rigidly adhere to its internal, pre-trained spatial priors even when they contradict the provided structural context.

**Visual Interference and Modality Conflict.** Comparing Text-Only with Multimodal GT SG settings reveals a counter-intuitive "visual interference" effect. Models often perform better without the image (e.g., LLaVA-OneVision-1.5 overall NAQ improves by +4.03% in the text-only setting), indicating that when symbolic context is perfect, noisy visual embeddings can act as distractors that confuse the logical engine rather than reinforcing the ground truth. In the more realistic Pred 3D SG setting, most models exhibit the modality trust bias, failing to self-correct when relying on their own imperfect perception. As illustrated in Fig. 5b, GPT 5 Mini's accuracy distribution remains stagnant, with a high rate of consistent hallucination (16.8%). A notable exception is Gemini 2.5 Flash, which utilizes its predicted graph as a cognitive scaffold, improving specifically in Counting (+6.92%) and Ranking (+2.51%) tasks by leveraging the intermediate structure to serialize complex enumeration.

**Case Study: Consistent Hallucination**

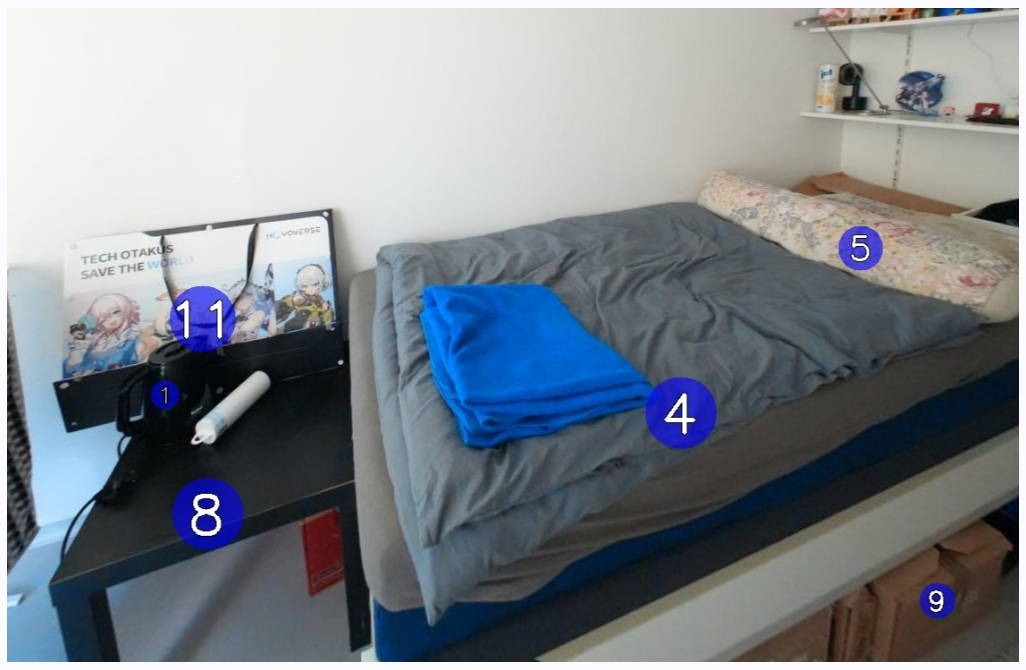


### QA Results

**Question:** From the camera's viewpoint, is object 1 in front of object 4?
A: Yes
B: No

**Ground Truth:** B
**Prediction:** A ✗
**Derived:** **A** ✗

### Scene Graph Snippet

```
...
"edges": [
    {
        "from": 4,
        "to": 1,
        "distance": 0.8,
        "relation": [
          "left",
          "in front of"
        ]
    },
    ...
```

**Diagnostic Insight:** Both the QA answer and the SGC prediction are factually incorrect. However, the model's textual reasoning strictly aligns with its flawed graph generation. This represents the consistent hallucination, indicating that the reasoning engine is tightly coupled with its wrong geometric grounding.

**Fig. 2:** A qualitative example of Qwen3-VL-8B demonstrating the consistent hallucination.

**Case Study: Imperfect Alignment**

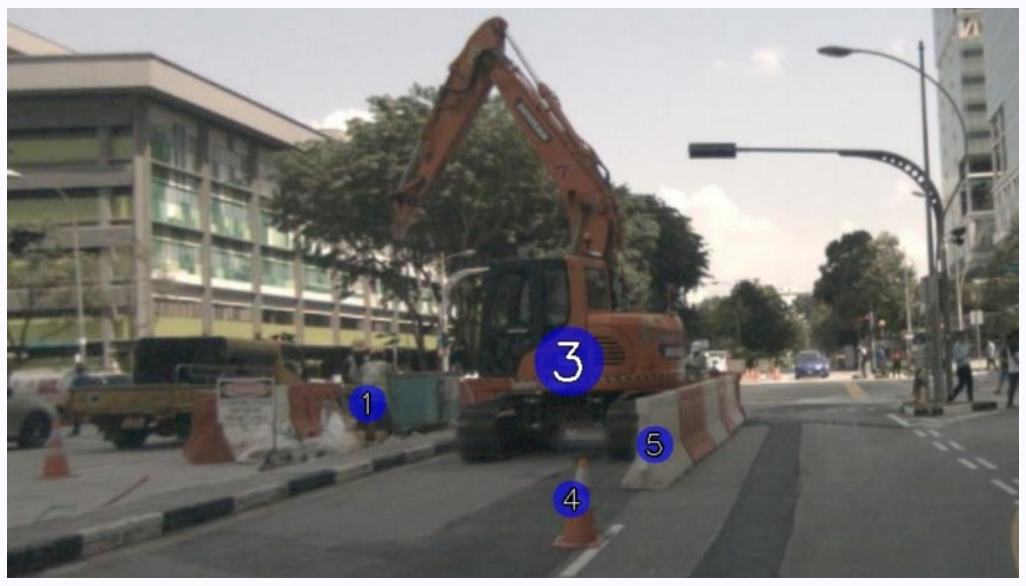


## QA Results

**Question:** Which object is physically closer to the camera: object 1 or object 5?
A: object 1
B: object 5

**Ground Truth:** B
**Prediction:** B ✓
**Derived:** B ✓

## Scene Graph Snippet

```
"objects": [
    {
        "id": 5,
        "dist_to_cam": 0.0,
        "size": {
          "w": 0.4,
          "l": 0.4,
          "h": 0.7
        }
    },
    {
        "id": 1,
        "dist_to_cam": 11.5,
        "size": {
          "w": 0.5,
          "l": 0.5,
          "h": 0.8
        }
    }
    ...
```

**Diagnostic Insight:** Although the QA answer is correct, a closer examination of the model's predicted SG reveals notable discrepancies with the ground truth. For instance, the predicted distance of Object 5 from the camera is 0, and the estimated height of Object 1 is roughly equivalent to that of Object 5. This represents imperfect alignment, a scenario wherein the model's structured perceptual representation is sufficient to deduce the correct QA response, yet its underlying numerical estimations deviate from physical reality.

**Fig. 3:** A qualitative example of Qwen3-VL-8B demonstrating the imperfect alignment.

**Case Study: Solver Failed**

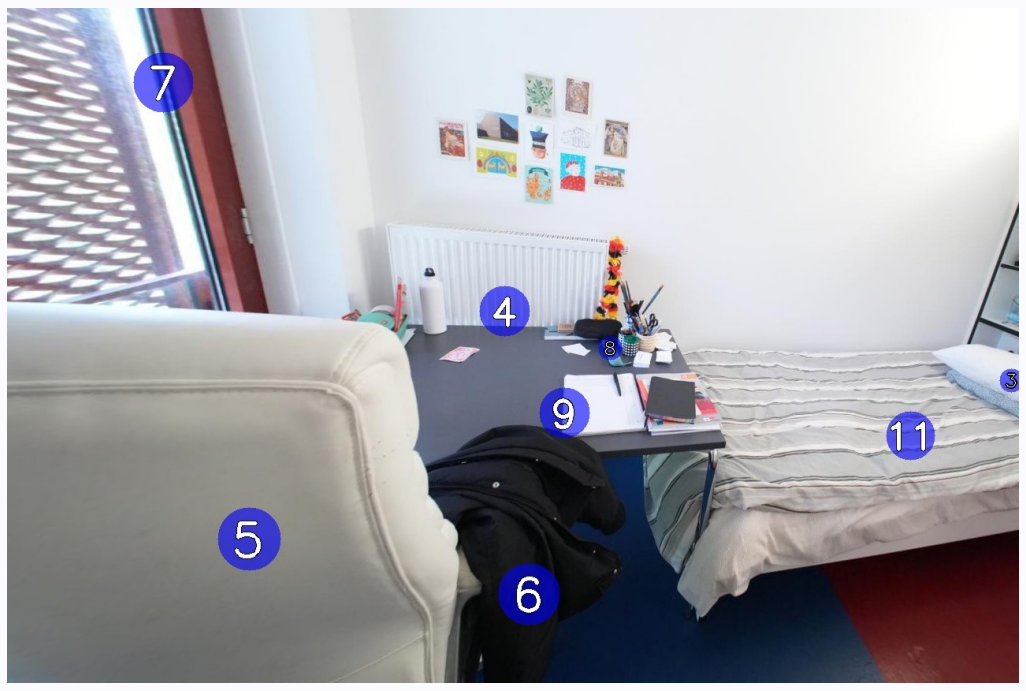


## QA Results

**Question:** Rank the following objects by height (tallest to shortest): object 7, object 5, object 6.
A: object 7-object 5-object 6
B: object 7-object 6-object 5
C: object 5-object 7-object 6
D: object 6-object 5-object 7

**Ground Truth:** A
**Prediction:** A ✓
**Derived:** [FAILED] ✗

## Scene Graph Snippet


```
"objects": [
    {
        "id": 5,
        "size": {
          "h": 0.2
        }
    },
    {
        "id": 6,
        "size": {
          "h": 0.15
        }
    },
    {
        "id": 7,
        "size": {
          "h": 0.05
        }
    }
    ...
```


**Diagnostic Insight:** In this specific instance, the predictions within the model's generated SG cannot be reconciled with any of the candidate answers; consequently, the programmatic solver outputs a `[FAILED]` token. This outcome denotes a solver failure, which is typically attributable to missing information within the predicted SG or an unresolvable mismatch with the available options, rather than a flaw in the solver's underlying execution logic.

**Fig. 4:** A qualitative example of LLaVA-OneVision1.5-8B demonstrating the solver failed.

**Table 6:** Performance comparison across different experimental settings. Bold denotes the best performance within each setting.

| | MCQ | | | | | | NAQ | | | |
|---|---|---|---|---|---|---|---|---|---|---|
| **Model** | **Counting** | **Direction** | **Distance** | **Ranking** | **Size** | **Transformation** | **Deduction** | **Distance** | **Size** | **Overall** |
| ***Setting: GT 3D SG intervention*** | | | | | | | | | | |
| Gemini-2.5-Flash | 85.37 | 70.74 | 98.04 | 86.37 | 92.51 | 28.28 | 46.98 | 99.41 | 94.15 | 77.99 |
| Gemini-3-Flash | **94.90** | **74.22** | **100.00** | 94.89 | 98.36 | **41.96** | **56.97** | **100.00** | **99.61** | **84.55** |
| GPT-5-Mini | 94.44 | 48.83 | **100.00** | **100.00** | **99.40** | 30.41 | 56.24 | **100.00** | 99.54 | 80.98 |
| GPT-5.2 | 80.16 | 68.20 | 99.17 | 80.81 | 94.23 | 28.00 | 49.33 | **100.00** | **99.61** | 77.72 |
| LLaVA-OV-1.5-8B | 58.84 | 69.67 | 79.03 | 60.45 | 75.82 | 21.63 | 50.58 | 99.53 | 73.71 | 65.47 |
| InternVL3.5-8B | 61.34 | 62.13 | 80.68 | 64.48 | 80.98 | 26.71 | 43.34 | 99.98 | 83.36 | 67.00 |
| Qwen2.5-VL-7B | 60.88 | 58.86 | 77.48 | 41.43 | 74.44 | 25.97 | 47.16 | 97.94 | 75.29 | 62.16 |
| Qwen3-VL-8B | 56.24 | 71.53 | 87.29 | 70.31 | 78.83 | 35.03 | 53.42 | 99.86 | 86.69 | 71.02 |
| ***Setting: Pred 3D SG intervention*** | | | | | | | | | | |
| Gemini-2.5-Flash | 65.08 | 62.13 | 75.00 | 53.72 | 70.22 | 21.26 | 41.11 | 34.78 | 49.69 | 52.55 |
| Gemini-3-Flash | **67.57** | **70.50** | **85.85** | **60.00** | **71.60** | **40.30** | **45.49** | 33.42 | **50.06** | **58.31** |
| GPT-5-Mini | 63.15 | 49.27 | 75.72 | 56.23 | 71.51 | 28.47 | 41.32 | 36.72 | 48.65 | 52.34 |
| GPT-5.2 | 63.49 | 65.17 | 80.37 | 55.16 | 68.67 | 27.45 | 40.58 | **42.88** | 49.29 | 54.79 |
| LLaVA-OV-1.5-8B | 56.01 | 63.01 | 65.19 | 34.62 | 61.96 | 21.90 | 41.44 | 27.68 | 25.63 | 44.16 |
| InternVL3.5-8B | 52.95 | 60.42 | 66.53 | 42.24 | 63.77 | 25.32 | 35.18 | 29.13 | 33.50 | 45.45 |
| Qwen2.5-VL-7B | 52.38 | 55.77 | 57.54 | 30.04 | 58.69 | 24.49 | 37.45 | 20.26 | 22.13 | 39.86 |
| Qwen3-VL-8B | 58.16 | 68.93 | 76.34 | 53.00 | 65.15 | 35.12 | 43.73 | 42.15 | 42.97 | 53.95 |
| ***Setting: GT 3D SG intervention without image input*** | | | | | | | | | | |
| Gemini-2.5-Flash | 78.46 | **69.72** | 96.28 | 87.00 | 91.22 | 28.74 | 40.76 | 99.58 | 89.15 | 75.66 |
| Gemini-3-Flash | 82.77 | 59.59 | 97.62 | 91.66 | 96.30 | 31.70 | 30.36 | **100.00** | 98.18 | 76.46 |
| GPT-5-Mini | **95.01** | 47.70 | **100.00** | **100.00** | **99.31** | **33.92** | **56.70** | **100.00** | **99.61** | **81.36** |
| GPT-5.2 | 73.47 | 61.30 | 98.04 | 80.99 | 93.29 | 32.62 | 46.53 | **100.00** | **99.61** | 76.21 |
| LLaVA-OV-1.5-8B | 60.54 | 61.40 | 83.88 | 69.87 | 84.25 | 23.38 | 46.08 | **100.00** | 96.13 | 69.50 |
| InternVL3.5-8B | 61.79 | 58.46 | 85.43 | 67.00 | 83.73 | 26.80 | 43.03 | **100.00** | 98.05 | 69.37 |
| Qwen2.5-VL-7B | 62.93 | 57.05 | 80.06 | 44.04 | 79.09 | 24.31 | 46.25 | 99.70 | 94.81 | 65.36 |
| Qwen3-VL-8B | 62.02 | 61.25 | 87.29 | 66.73 | 79.35 | 28.10 | 47.76 | 99.77 | 98.38 | 70.07 |

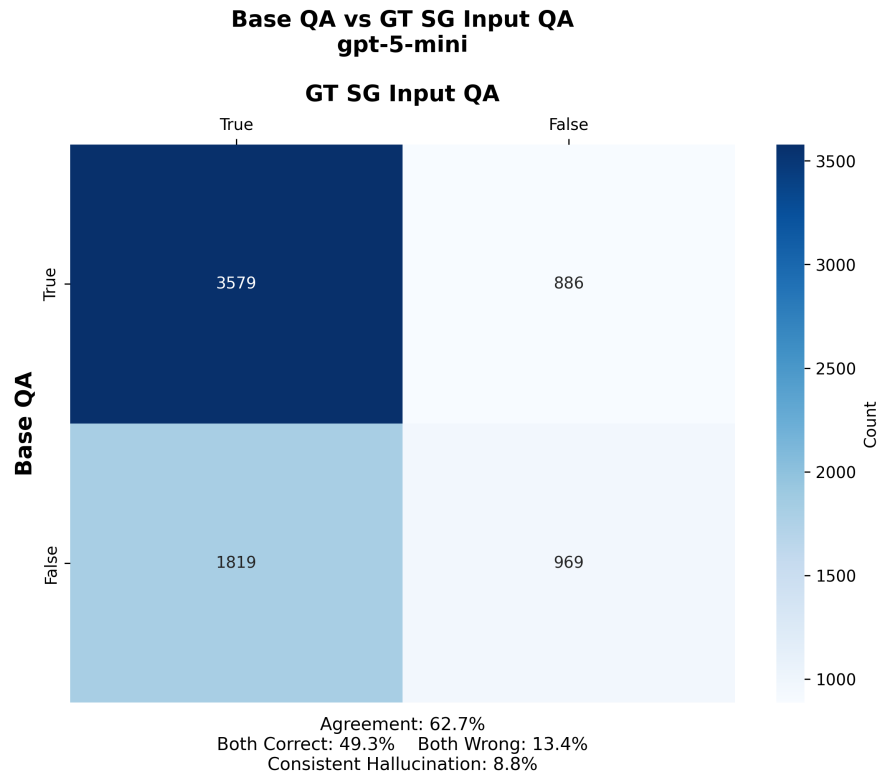


**(a)** Base QA vs GT SG Input QA

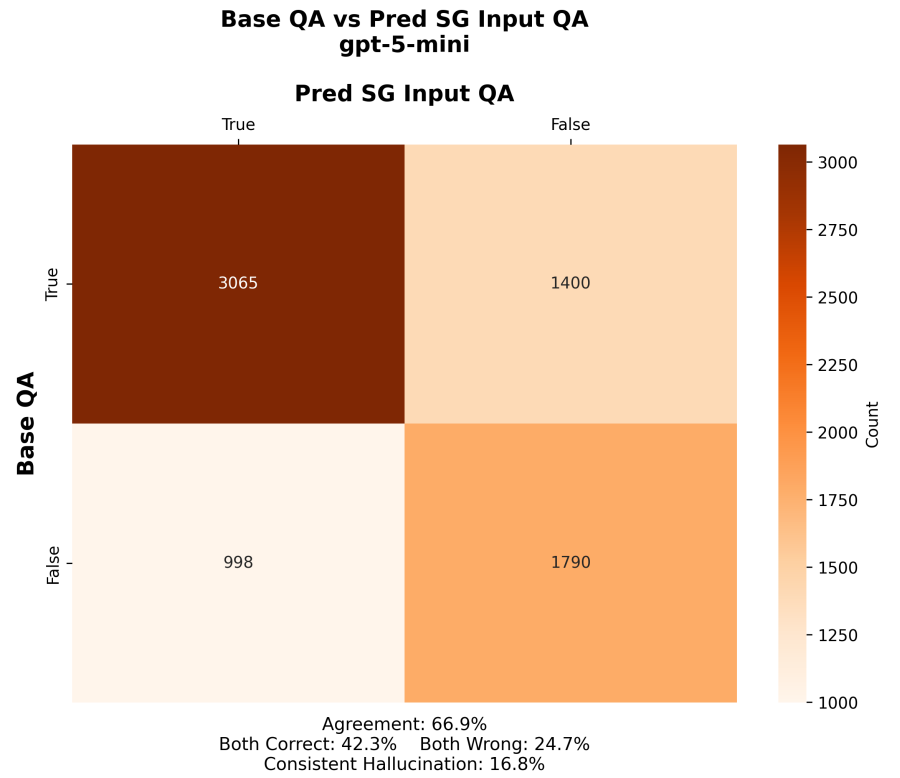


**(b)** Base QA vs Pred SG Input QA

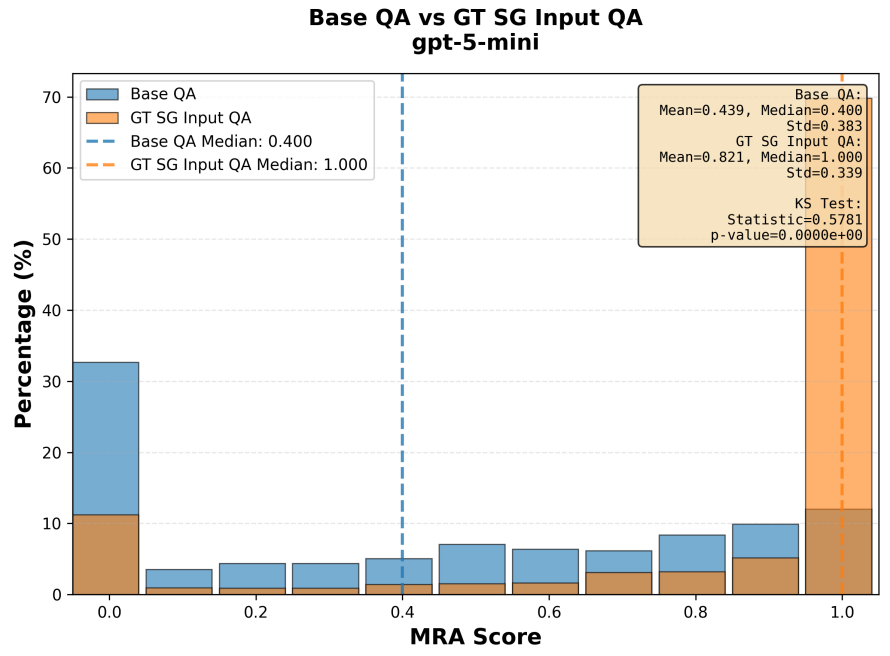


**(c)** Base QA vs GT SG Input QA

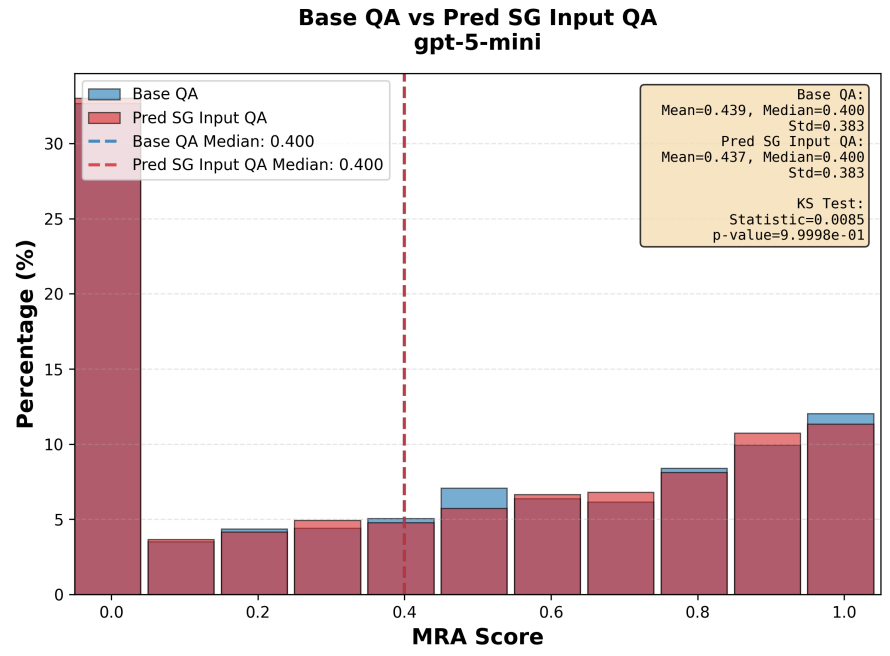


**(d)** Base QA vs Pred SG Input QA

**Fig. 5: Case Study on GPT-5-Mini.** Under GT 3D SG intervention, the model corrects 1,819 previously failed queries and achieves near-perfect metric estimation (Median MRA 1.0). Under Pred 3D SG intervention, the metric distribution stagnates (Median MRA 0.4), and the model exhibits high consistent hallucination, confirming that internal perceptual bottlenecks prevent the reasoning engine from activating.

### B.7 Diagnostic Insight: Relation Confusion Matrix

To scrutinize the intrinsic spatial modeling capabilities of VLMs, especially the relation-capture ability, we analyze the edge distributions within the generated 3D Scene Graphs (SGC task). We categorize all spatial edges into three orthogonal groups: Front-Back (Depth), Left-Right (Lateral), and Up-Down (Vertical). In this analysis, "none" denotes the absence of a spatial relation in a specific dimension; for instance, if the ground truth defines an edge strictly as "left," the corresponding labels for the Front-Back and Up-Down dimensions are considered "none." Fig. 6 illustrates the confusion matrices for representative proprietary and open-source models.

A critical divergence is observed in the vertical dimension, where open-source models exhibit a systematic *vertical neglect*. While proprietary models like GPT 5 Mini and Gemini 2.5 Flash demonstrate robust omnidirectional awareness with high recall for Up-Down relations, open-source models such as LLaVA-OneVision-1.5 and Qwen3-VL overwhelmingly predict "none" even when vertical relations explicitly exist in the ground truth. This suggests that the visual encoders of open-source models are heavily biased towards horizontal, ground-plane navigation, effectively collapsing the 3D perceptual field into a 2D representation, whereas proprietary models maintain a more holistic 3D spatial awareness.

Furthermore, a universal hierarchy of difficulty is evident across all architectures: lateral stability vs. depth ambiguity. The Left-Right relations consistently achieve the highest accuracy and diagonal density across all models, confirming that 2D lateral layout perception is well-established. Conversely, performance degrades significantly in the Front-Back dimension. While proprietary models manage to capture depth cues with reasonable success, open-source models struggle significantly, often defaulting to "none" or exhibiting lower recall. This *depth blindness*, combined with the aforementioned vertical neglect, highlights that while current open-source VLMs excel at 2D semantic recognition, they lack the fundamental Z-axis and vertical geometric grounding required for true embodied intelligence.

## C Validation of Experimental Protocol Design

In this section, we present supplementary ablation studies to validate the rigor and effectiveness of the experimental protocols adopted in the Sec. 4. To ensure computational feasibility while maintaining diagnostic resolution, these experiments are conducted on a subset of 100 scenes randomly sampled from the CRISP benchmark, maintaining a strict 1:1 indoor-outdoor ratio to minimize domain bias. This subset comprises a total of 862 QA pairs. Our analysis focuses on justifying two pivotal design choices: the structural generation format and the inference strategy.

First, we verify the efficacy of our object-centric star-topology generation by contrasting it with a pairwise baseline, demonstrating that holistic context is essential for structural fidelity (Sec. C.1). Second, we investigate the impact of test-time compute (e.g., Chain-of-Thought [11] and "thinking" modes). We provide

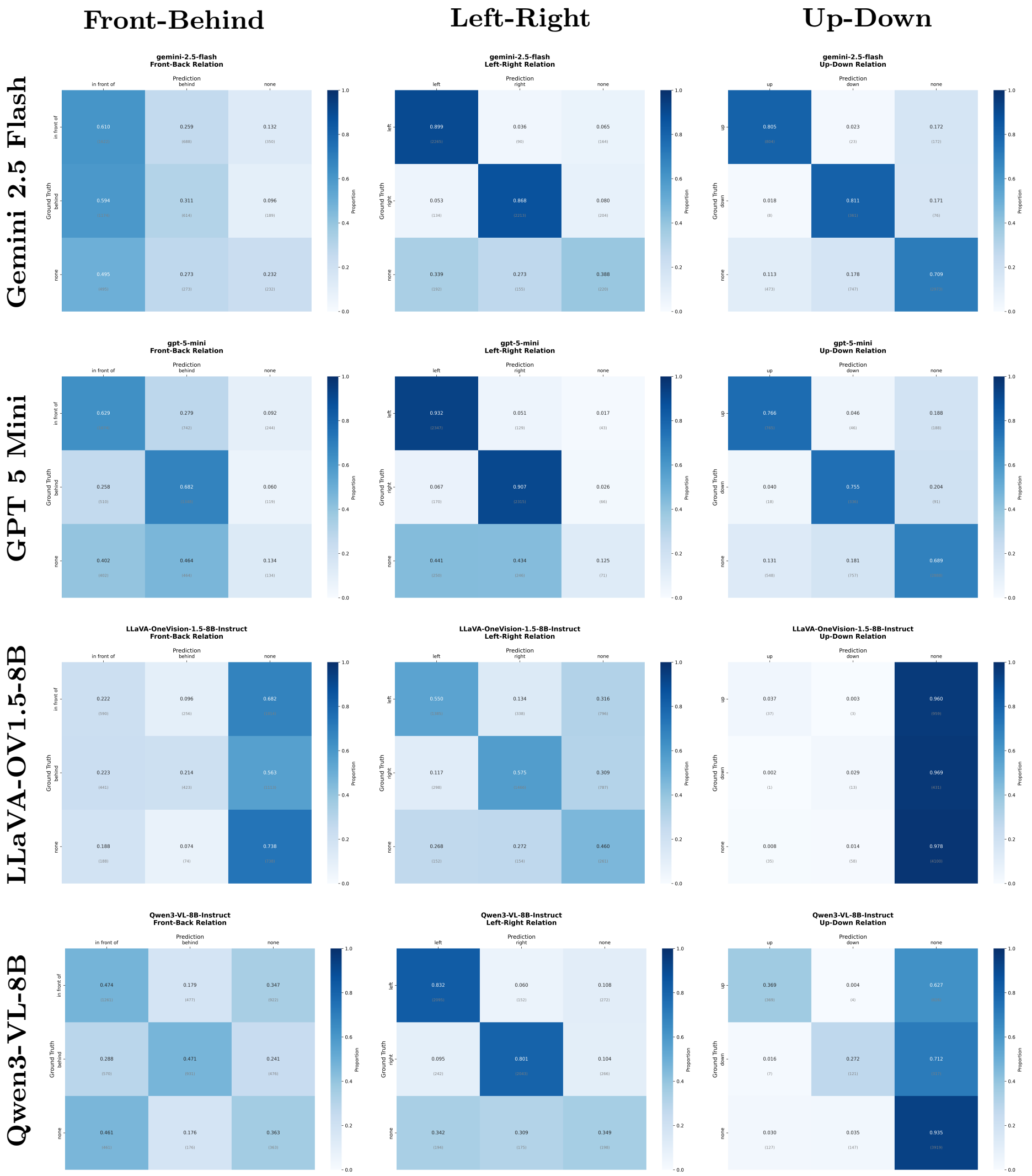


**Fig. 6: Confusion matrices of generated spatial relations across three orthogonal dimensions.** We compare the structural generation performance of two proprietary models (Gemini 2.5 Flash, GPT 5 Mini) and two open-source models (LLaVA-OneVision-1.5-8B, Qwen3-VL-8B). The label “none” indicates the absence of a relation in a specific dimension within either the ground truth or the model’s prediction.

empirical evidence that increasing the reasoning budget yields disproportionate gains in QA performance compared to consistency. This widening gap confirms that "thinking" primarily amplifies semantic shortcuts rather than resolving the fundamental metric grounding bottleneck, thereby justifying our decision to restrict these features to isolate intrinsic spatial intelligence (Sec. C.2).

### C.1 Holistic Generation vs. Pairwise Assembly

**Table 7:** Evaluation results of Pairwise SGC ablation. The top section shows results for the Pairwise setting, and the bottom section shows the Holistic setting. **Bold** denotes the best performance.

| Model | Size | Cam Dist | Dist | Est | Rel | Overall | Comp Rate |
|---|---|---|---|---|---|---|---|
| ***Setting: Pairwise*** | | | | | | | |
| Gemini-2.5-Flash | **69.47** | 56.80 | **56.70** | **60.99** | **64.79** | **62.89** | **100.00** |
| LLaVA-OV-1.5-8B | 42.77 | 43.24 | 40.50 | 42.17 | 55.93 | 49.05 | **100.00** |
| Qwen2.5-VL-7B | 56.43 | 47.44 | 36.70 | 46.86 | 52.61 | 49.74 | **100.00** |
| Qwen3-VL-8B | 61.30 | **65.71** | 51.71 | 59.57 | 58.79 | 59.18 | 99.45 |
| ***Setting: Holistic*** | | | | | | | |
| Gemini-2.5-Flash | **79.21** | **60.79** | **53.35** | **64.45** | **77.62** | **71.03** | **100.00** |
| LLaVA-OV-1.5-8B | 42.85 | 34.47 | 38.74 | 38.69 | 65.63 | 52.16 | **100.00** |
| Qwen2.5-VL-7B | 50.92 | 28.83 | 32.36 | 37.37 | 53.91 | 45.64 | 98.00 |
| Qwen3-VL-8B | 64.83 | 45.19 | 46.70 | 52.24 | 73.90 | 63.07 | 98.00 |

To validate the architectural choice of our Object-Centric Star-Topology generation (Sec. 3.2), we compare our standard *Holistic* SGC approach against a *Pairwise* SGC baseline. In the Holistic setting, the model is required to output the complete scene graph, encoding the center object and all $N$ target neighbors in a single inference pass. Conversely, the Pairwise ablation decomposes this task into $N$ independent inference steps, each targeting the center object and a single neighbor. To ensure a strictly fair comparison, we aggregate the pairwise outputs into a unified graph structure. While edge predictions are unique to each pair, the center node is predicted $N$ times; thus, we calculate the arithmetic mean of the metric scores (Size and Distance-to-Camera) across all predictions for the center node to derive the final node accuracy.

Tab. 7 reveals that three out of four models achieve higher overall SGC Scores in the Holistic setting. This result is driven by two competing cognitive mechanisms:

1. **Local Attention vs. Global Anchoring:** The Pairwise setting acts as a strong attention guide, forcing the model to focus exclusively on the visual features of two specific objects. As observed with Qwen2.5-VL-7B, this local

focus can enhance metric precision ($S_{est}$) by reducing background distraction. However, this comes at the cost of "global anchoring." In the Holistic setting, the simultaneous visibility of multiple objects allows the model to triangulate positions. The significant boost in relation scores ($S_{rel}$) for Gemini 2.5 Flash (+12.83%) and Qwen3-VL-8B (+15.11%) confirms that this global context is indispensable for constructing a coherent topological structure, which outweighs the minor gains in local metric precision.

2. **Sensitivity of Spatial Metrics:** A potential concern is whether the overall improvement is skewed by the high volatility of the relation score. We argue that this disparity highlights a fundamental characteristic of VLMs: metric estimation ($S_{est}$) is largely driven by object-level semantic priors (e.g., recognizing a "chair" and retrieving its typical size), which are robust to context length. In contrast, spatial relations ($S_{rel}$) are contextual. The drop in $S_{rel}$ under the Pairwise setting (e.g., LLaVA-OneVision-1.5: 65.63% $\rightarrow$ 55.93%) exposes that without holistic context, models lose the structural "skeleton" of the scene. Since true spatial intelligence requires both accurate metrics and correct topology, the Holistic approach provides the necessary information density to maximize the joint probability of both.

Finally, despite the increased sequence length, all models maintain a near-perfect syntactic compliance rate (>98%) in the Holistic setting (Tab. 7). This confirms that generating complex, holistic scene graphs does not induce an execution bottleneck, validating our protocol as both cognitively superior and computationally feasible.

### C.2 Impact of Reasoning Budget and Chain-of-Thought

To rigorously justify our decision to restrict test-time compute in the primary evaluations, we investigate how scaling the inference reasoning budget impacts structural spatial perception. We contrast two experimental configurations: the *Baseline* (adopted in Sec 4.1) and the *Extended Reasoning* setting. In the Baseline, standard models receive no explicit reasoning prompts, and specialized reasoning models are constrained to a minimal budget to prevent extended self-correction (e.g., Gemini 2.5 Pro is limited to a thinking budget of 1,024 tokens; GPT 5 Mini is set to "low" reasoning effort). Conversely, the Extended Reasoning setting prompts standard models with a zero-shot CoT ("Let's think step by step"), and significantly amplifies the compute allocated

**Table 8:** Effect of extended reasoning. We report the aggregate QA, SGC, Derived QA (Der. QA), and Consistency (Con.) scores. **Bold** denotes the best performance within each category.

| Model | QA | SGC | Der. QA | Con. |
|---|---|---|---|---|
| ***Setting: Extended Reasoning*** | | | | |
| Gemini-2.5-Flash* | 52.95 | 63.80 | 58.29 | 53.88 |
| Gemini-2.5-Pro† | **61.42** | 63.90 | 57.74 | 60.46 |
| GPT-5-Mini† | 59.88 | **64.54** | **58.36** | **62.63** |
| Qwen3-VL-8B* | 52.31 | 60.95 | 54.05 | 47.24 |
| ***Setting: Baseline*** | | | | |
| Gemini-2.5-Flash | 47.74 | 63.51 | **58.35** | 52.93 |
| Gemini-2.5-Pro | **59.43** | 63.80 | 57.85 | 58.39 |
| GPT-5-Mini | 55.28 | **66.35** | 56.48 | **60.80** |
| Qwen3-VL-8B | 57.11 | 62.22 | 53.55 | 52.24 |

†: Denotes models utilizing native extended thinking modes. *: Denotes models prompted with Zero-shot CoT.

to proprietary models (e.g., configuring Gemini 2.5 Pro with a 16,384-token thinking budget and setting GPT 5 Mini to "high" reasoning effort). The results, detailed in Tab. 8, expose a critical divergence between linguistic reasoning output and physical metric grounding.

A detailed analysis of the proprietary reasoning models reveals the deceptive nature of extended compute in spatial tasks. When allocated a high reasoning budget, both Gemini 2.5 Pro and GPT 5 Mini exhibit noticeable gains in QA accuracy and a marginal improvement in their Consistency scores. However, this is sharply contradicted by their SGC performance, which remains stagnant or actively degrades (e.g., GPT 5 Mini's SGC score drops from 66.35% to 64.54%). This paired phenomenon unveils a subtle cognitive mechanism: while extended reasoning may enhance the model's ability to mobilize its implicit spatial modeling, it simultaneously amplifies the dominant influence of language priors. During the prolonged reasoning process, the model constructs a strong linguistic narrative of the scene. When subsequently forced to generate the 3D Scene Graph, these entrenched linguistic priors frequently override the raw perceptual signals from the vision encoder. Consequently, the model artificially aligns its structural output with its hallucinated text (explaining the slight rise in Consistency) without actually improving its grounding in the physical image, leading to a stagnated SGC score.

Crucially, this finding of open-source models collapse while proprietary models show superficial QA gains aligns with recent literature observing the limited efficacy of CoT in VLM's spatial problem-solving [7,8,12,14]. However, our granular structural diagnosis provides a deeper mechanistic explanation for these prior findings. The perceived "improvement" in proprietary models is largely a linguistic illusion driven by the aforementioned semantic overwriting. Meanwhile, for open-source architectures like Qwen3-VL-8B, the application of zero-shot CoT introduces severe structural instability, causing simultaneous collapses in QA, SGC, and Consistency scores. This occurs because forcing open-source VLMs to generate lengthy, unstructured spatial rationales disrupts their delicate cross-modal alignment, leading to compounded hallucination rather than logical deduction. Ultimately, extended reasoning fails to resolve the fundamental visual grounding bottleneck across all architectures. This validates our Baseline protocol as a necessary constraint to strictly isolate and evaluate the models' immediate, intrinsic geometric perception without the confounding effects of post-hoc linguistic rationalization.

## D Roadmap for Architectural Intervention

To maximize the practical utility of the CRISP benchmark, researchers can utilize the roadmap (Fig. 7) in a systematic, two-step debugging process. First, models are profiled under the default setting using Panel A (Primary Diagnosis). By navigating the decision nodes based on Consistency, QA, and SGC scores, researchers can rapidly identify the macroscopic failure mode. For example, navigating this panel immediately reveals whether a model's failure stems from

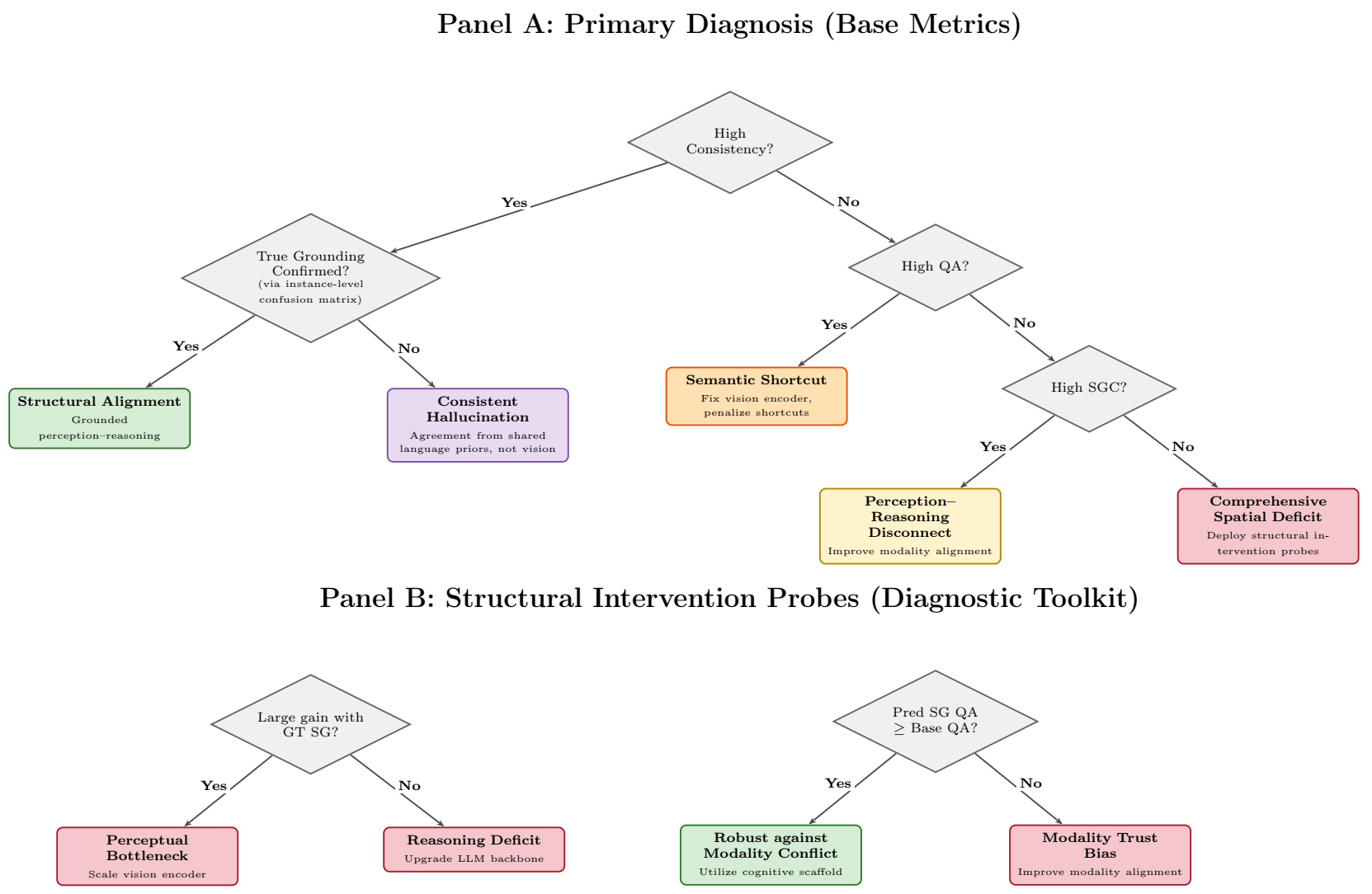


**Fig. 7: The CRISP Diagnostic Roadmap.** A modular framework for suggesting potential architectural intervention. Panel A leverages baseline metrics to diagnose primary phenotypic failure modes. Panel B provides independent structural probes to mechanistically decouple perceptual bottlenecks from reasoning deficits and evaluate robustness against modality conflict.

ignoring geometry entirely (diagnosed as a *Semantic Shortcut* when a disproportionately high QA score is coupled with a low SGC or Consistency score) or failing to leverage the geometry it successfully perceived (*Perception-Reasoning Disconnect*). Crucially, high consistency must be further validated via instance-level confusion matrices to rule out *Consistent Hallucination* driven by language priors.

Second, to translate these macroscopic symptoms into hypothetical engineering solutions, researchers can utilize the independent tools in Panel B (Intervention Probes). These probes are designed to isolate entangled capabilities. For instance, if Panel A diagnoses a *Comprehensive Spatial Deficit*, the *Ground-Truth SG probe* (Ceiling Test) determines whether compute might be better allocated to scaling the vision encoder (*Perceptual Bottleneck*) or upgrading the LLM backbone (*Reasoning Deficit*). Similarly, for models struggling with alignment, the *Predicted SG probe* (Robustness Test) isolates whether the failure is exacerbated by a *Modality Trust Bias*, a critical flaw where models prioritize flawed self-generated text over raw visual evidence. By decoupling the diagnosis in this manner, CRISP aims to guide researchers in formulating more precisely targeted hypotheses for addressing the underlying cognitive bottleneck.

Finally, to ensure the longevity and adaptability of this roadmap, the diagnostic thresholds (e.g., "High" and "Low") must be interpreted as relative anchors rather than absolute numerical cutoffs. Given the nascent state of spa-

tial reasoning in current VLMs, a metric should be categorized as "Low" if it does not meaningfully surpass the text-only blind baseline or significantly trails behind the cohort average, indicating a fundamental lack of visual grounding. Conversely, categorical decisions rely heavily on cross-model benchmarking and intra-model metric divergence. For instance, a QA score is considered "High" if it ranks near the state-of-the-art within the evaluated cohort, yet simultaneously exhibits an asymmetry relative to its own lagging SGC score, exposing a clear cognitive imbalance. By grounding these decision boundaries in cohort-relative performance and cross-metric disparities, this diagnostic framework will remain robust and highly relevant even as absolute multimodal capabilities advance in future model generations.

## E Ethics and Accessibility

**Data Ethics and Privacy.** The CRISP benchmark is constructed upon two well-established, publicly available datasets: nuScenes for macro-scale outdoor driving environments and ScanNet++ for micro-scale indoor scenes. We strictly adhere to their respective academic licenses and terms of use. Crucially, the data curated in our benchmark is utilized exclusively for scientific evaluation and diagnostic purposes, rather than model training. This strict evaluation-only protocol inherently mitigates the risk of downstream models memorizing and inadvertently leaking residual information. Furthermore, we rely on the rigorous privacy-preserving measures implemented by the original dataset creators, such as the anonymization of Personally Identifiable Information (PII) in nuScenes and the informed consent protocols employed during the collection of ScanNet++. We have manually reviewed the sampled subsets to ensure the absence of offensive content or sensitive PII, ensuring that all figures and illustrations in this manuscript comply with strict ethical standards.

**Potential Societal Impacts.** By focusing specifically on the evaluation of visual spatial intelligence in indoor and outdoor environments, our research aims to accelerate the safe deployment of embodied AI. The positive societal impact of CRISP lies in its ability to expose the "perception-reasoning disconnect," guiding the community away from brittle, hallucination-prone models toward physically grounded systems. This is a critical prerequisite for safety-critical applications such as autonomous driving, robotic caregiving, and navigation aids for the visually impaired. Conversely, the primary negative societal risk stems not from the malicious use of our benchmark, but from the premature integration of current VLMs into these physical environments. If developers over-rely on models that excel in standard linguistic QA but fail in metric spatial grounding, it could lead to severe physical accidents in domestic or vehicular settings. Additionally, deploying spatially-aware models in indoor domestic spaces inherently touches upon household privacy, necessitating highly reliable and predictable model behaviors.

**Mitigation and Accessibility.** To mitigate these risks of premature deployment and to foster collaborative progress, we are fully committed to the open-

access dissemination of our diagnostic tool. In strict accordance with the reproducibility and open-source policies of the computer vision community, we commit to publicly releasing the complete CRISP benchmark prior to the camera-ready deadline, contingent upon acceptance. Crucially, to rigorously respect the end-user license agreements and copyright constraints of the source datasets (specifically the strict prohibition on raw data redistribution mandated by the ScanNet++), we adopt an annotation-only release protocol. We will open-source our complete set of annotations (including the Master 3D Scene Graphs and the 9,839 QA pairs) and the cross-task consistency evaluation codebase under an appropriate academic license. For the raw visual data, we will provide an automated setup pipeline. Users will be required to independently obtain the original images from the respective data providers after agreeing to their specific terms.



\end{document}